\documentclass[]{article}

\usepackage{booktabs} % For formal tables
\usepackage{cite}
\usepackage{microtype}
\usepackage{xspace}
\usepackage{hyperref}
\usepackage{array}
\usepackage{dsfont}
\usepackage{pifont}
\usepackage{helvet}
\usepackage{courier}
\usepackage{color}
\usepackage{colortbl}
\usepackage{mathrsfs}
\usepackage{amscd}
\usepackage{stmaryrd}
\usepackage{amssymb}
\usepackage{amsmath}
\usepackage{graphicx}
\usepackage{algorithm} 
\usepackage[noend]{algpseudocode}
\algrenewcommand\algorithmicindent{0.750em}%
\usepackage{subfig}
\usepackage{multirow}
\usepackage{tikz}
\usetikzlibrary{shapes,arrows}
\usepackage{authblk}

\newcommand{\commentout}[1]{}

\newcommand{\networks}{\mathcal{N}}

\newcommand{\testsuites}{\mathcal{T}}

\newcommand{\distance}[2]{ ||#1||_{#2}}

\newcommand{\distancefunction}{h}
\newcommand{\neuronpair}{\alpha}

\newcommand{\metric}{M}

\newcommand{\transform}{\delta}

\newcommand{\DRL}{DR}
\newcommand{\DRLp}{DR$^+$}

\newcounter{point}
\newtheorem{definition}{Definition}

\begin{document}

\title{Concolic Testing for Deep Neural Networks
\thanks{Kwiatkowska and
Ruan are supported by EPSRC Mobile Autonomy Programme Grant (EP/M019918/1).
Wu is supported by the CSC-PAG Oxford Scholarship. 
}} 

\author[1]{Youcheng Sun}
\author[1]{Min Wu}
\author[1]{Wenjie Ruan}
\author[2]{Xiaowei Huang}
\author[1]{Marta Kwiatkowska}
\author[1]{Daniel Kroening}

\affil[1]{University of Oxford, UK}
\affil[ ]{\texttt {\normalsize \{youcheng.sun; min.wu; wenjie.ruan\}@cs.ox.ac.uk}}
\affil[ ]{\texttt {\normalsize \{marta.kwiatkowska; daniel.kroening\}@cs.ox.ac.uk}}

\affil[2]{University of Liverpool, UK}
\affil[ ]{\texttt {\normalsize xiaowei.huang@liverpool.ac.uk}}

\date{}

\maketitle

\begin{abstract}
Concolic testing combines program execution and symbolic analysis to explore the
execution paths of a software program. This paper presents the first concolic
testing approach for Deep Neural Networks (DNNs). More specifically, we formalise
coverage criteria for DNNs that have been studied in the literature, and then develop a
coherent method for performing concolic testing to increase test coverage.
Our experimental results show the effectiveness of the concolic testing
approach in both achieving high coverage and finding adversarial examples.
\end{abstract}

%\keywords{neural networks, symbolic execution, test case generation}

\newcommand\true{\textsf{true}\xspace}
\newcommand\false{\textsf{false}\xspace}

\section{Introduction}
\label{sec:intro}

Deep neural networks (DNNs) have been instrumental in solving a range of hard problems in AI, e.g., the ancient game of Go, image classification, and natural language processing. As a result, many potential applications are envisaged.
%and have been applied to many applications.  
%
%When it is applied to very high dimensional problems such as the classification of \emph{natural} images, which have e.g., 299*299*3 dimensions in the imageNet dataset \cite{}, the training dataset, even with millions of inputs, forms a small, or negligible, proportion of all possible inputs. 
%
%Therefore, before its application to safety critical systems on which human lives are at stake, serious verification or testing should be held to ensure that a DNN's predictions on those inputs that are not in the training dataset are correct. 
%
% can be applied to other inputs that are not in its training dataset (i.e., generalisability problem). 
%
However, major concerns have been raised about the suitability of this technique 
for safety- and security-critical systems, where faulty behaviour carries the risk of endangering human lives or financial damage. To address these concerns, a (safety or security) critical system comprising DNN-based components needs to be validated thoroughly.

The software industry relies on testing as a primary means to provide stakeholders with information about the quality of the software product or service under test \cite{Kaner2006}.  
So far, there have been only few attempts to test DNNs systematically~\cite{WHK2018, deepxplore, tian2017deeptest, SHK2018,deepgauge}. These are either based on concrete execution, e.g., Monte Carlo tree search~\cite{WHK2018} or gradient-based search~\cite{deepxplore,tian2017deeptest,deepgauge}, or symbolic execution in combination with solvers for linear arithmetic~\cite{SHK2018}. Together with these test-input generation algorithms, several test coverage criteria have been presented, including neuron coverage~\cite{deepxplore},
a criterion that is inspired by MC/DC~\cite{SHK2018}, and criteria to capture particular neuron activation values to identify corner cases~\cite{deepgauge}. None of these approaches implement \emph{concolic testing}~\cite{dart, sen2005cute}, which combines concrete execution and symbolic analysis to explore the execution paths of a program that are hard to cover by
techniques such as random testing.

We hypothesise that concolic testing is particularly well-suited for DNNs. The input space of a DNN is usually high dimensional, which makes random testing difficult. For instance, a DNN for image classification takes tens of thousands of pixels as input. Moreover, owing to the widespread use of the ReLU activation function for hidden neurons, the number of ``execution paths" in a DNN is simply too large to be completely covered by symbolic execution. Concolic testing can mitigate this
complexity by directing the symbolic analysis to particular execution paths, through concretely evaluating given properties of the DNN.

In this paper, we present the first concolic testing method for DNNs. The method is parameterised using a set of coverage requirements, which we express using Quantified Linear Arithmetic over Rationals (QLAR). For a given set $\mathfrak{R}$ of coverage requirements, we incrementally generate a set of test inputs to improve coverage by alternating between concrete execution and symbolic analysis. Given an unsatisfied test requirement~$r$, we identify a test input $t$ within our current test suite such that $t$ is close to satisfying $r$ according to an evaluation based on \emph{concrete execution}. After that, \emph{symbolic analysis} is applied to obtain a new test input~$t'$ that satisfies~$r$. The test input $t'$ is then added to the test suite. This process is iterated until we reach a satisfactory level of coverage.

Finally, the generated test suite is passed to a \emph{robustness oracle}, which
determines whether the test suite includes \emph{adversarial examples}~\cite{SZSBEGF2014}, i.e., 
pairs of test cases that disagree on their classification labels when close to each other with respect to a given distance metric. %can be both 
%close to any of the examples with respect to a distance measure and have the same classification label, in which case the test passes the oracle. % with it. 
%We call this oracle a robustness oracle. 
The lack of robustness has been viewed as a major weakness of DNNs, and the discovery of adversarial examples and the robustness problem are studied actively in several domains, including machine learning, automated verification, cyber security, and software testing. 

\noindent
Overall, the main contributions of this paper are threefold:

\begin{enumerate}
    \item We develop the first concolic testing method for DNNs.
    \item We evaluate the method with a broad range of test coverage requirements, including Lipschitz continuity~\cite{WB2015, ACB2017, WHK2018, BSZ2017, RHK2018}
    and several structural coverage metrics~\cite{deepxplore, SHK2018, deepgauge}. We show experimentally that our new algorithm supports this broad range of properties in a coherent way.
    \item We implement the concolic testing method in the software tool \emph{DeepConcolic}\footnote{ \url{https://github.com/TrustAI/DeepConcolic}}.
    Experimental results show that DeepConcolic achieves high coverage and that it is able to discover a significant number of adversarial examples.
\end{enumerate}

%Concolic testing \cite{} combines the strength of symbolic analysis and
%concrete executions, and has been shown to be able to improve code coverage
%for its ability to automatically generate test cases to explore execution
%paths in a program that are hard to be explored by either symbolic analysis
%or concrete executions themselves.  In this paper, we leverage the
%capability of concolic testing in working with a new subject of study for
%software testing: DNNs.

\section{Related Work}
\label{sec:related}

We briefly review existing efforts for assessing the robustness of DNNs
and the state of the art in concolic testing.

\subsection{Robustness of DNNs}

Current work on the robustness of DNNs can be categorised as offensive or
defensive.  Offensive approaches focus on heuristic search algorithms
(mainly guided by the forward gradient or cost gradient of the DNN) to find
adversarial examples that are as close as possible to a correctly classified
input.  On the other hand, the goal of defensive work is to increase the
robustness of DNNs.  There is an arms race between offensive and defensive
techniques.

In this paper we  focus on defensive methods. 
A promising approach is automated verification, which aims to provide robustness guarantees for DNNs.
The main relevant techniques include a layer-by-layer exhaustive search~\cite{HKWW2017}, methods that use
constraint solvers~\cite{KBDJK2017}, global optimisation approaches~\cite{RHK2018} and abstract interpretation~\cite{gehr2018ai, mirman2018differentiable}
to over-approximate a DNN's behavior. Exhaustive search suffers from the state-space explosion problem, which can be alleviated by Monte Carlo tree search~\cite{WHK2018}.
Constraint-based approaches are limited to small DNNs
with hundreds of  neurons.  Global optimisation improves over
constraint-based approaches through its ability to work with large DNNs, but
its capacity is sensitive to the number of input dimensions that need to be
perturbed. The results of over-approximating analyses can be pessimistic because of
false alarms.

The application of traditional testing techniques to DNNs is difficult, and
work that attempts to do so is more recent, e.g.,~\cite{WHK2018,
deepxplore, tian2017deeptest, SHK2018, deepgauge}.  Methods inspired
by software testing methodologies typically employ coverage criteria to
guide the generation of test cases; the resulting test suite is then
searched for adversarial examples by querying an oracle.  The coverage
criteria considered include \emph{neuron coverage}~\cite{deepxplore},
which resembles traditional statement coverage.  A~set of criteria inspired
by MD/DC coverage~\cite{HVCR2001} is used in~\cite{SHK2018};
Ma et al.~\cite{deepgauge} present criteria that are designed to capture particular
values of neuron activations. 
Tian et al.~\cite{tian2017deeptest} study the utility of neuron coverage for detecting adversarial examples in DNNs for the Udacity-Didi Self-Driving Car Challenge.

We now discuss algorithms for test input generation. 
Wicker et al.~\cite{WHK2018} aim to cover the input space by exhaustive
mutation testing that has theoretical guarantees, while in~\cite{deepxplore, tian2017deeptest, deepgauge}
gradient-based search algorithms are applied to solve optimisation problems, and Sun et al.~\cite{SHK2018}
apply linear programming. None of these consider concolic testing and a general means for 
modeling test coverage requirements as we do in this paper. 

\subsection{Concolic Testing}

By concretely executing the program with particular inputs, which includes random testing, 
a large number of inputs can be tested at low cost. However, without guidance, the generated test cases may be restricted to a subset of the execution paths of the program and the probability of exploring 
execution paths that contain bugs can be extremely low.
In symbolic execution~\cite{cadar2008klee, xie2005symstra, visser2004test}, an execution
path is encoded symbolically. Modern constraint solvers can determine feasibility of the encoding effectively, although performance still degrades as the size of the symbolic representation increases.
Concolic testing~\cite{dart, sen2005cute} is an effective approach to automated test input generation. It is a hybrid software testing technique that alternates between concrete execution, i.e., testing on particular inputs, and symbolic execution, a classical technique that treats program variables as symbolic values~\cite{KHC2015}.

Concolic testing has been applied routinely in software testing, and a wide range of tools 
is available, e.g.,~\cite{dart, sen2005cute, burnim2008heuristics}. 
It starts by executing the program with a concrete input.
At the end of the concrete run, another execution path must be selected heuristically. 
This new execution path is then encoded symbolically and the resulting formula is solved by a constraint solver, to yield a new concrete input. The concrete execution and the symbolic analysis alternate until a desired level of structural coverage is reached.

The key factor that affects the performance of concolic testing is the heuristics used to select the next execution path. While there are simple approaches such as random search and depth-first search,  
more carefully designed heuristics can achieve better coverage~\cite{burnim2008heuristics, godefroid2008automated}.
Automated generation of search heuristics for concolic testing is an active area of research~\cite{lin2018, cha2018}.

\begin{table}[ht]
    \caption{Comparison with different coverage-driven DNN testing methods}
    \label{tab:testings}
    \centering
    \scalebox{0.58}{
    \def\arraystretch{1.2}
    \begin{tabular}{l||c|c|c|c|c}
    \toprule
         & \textbf{DeepConcolic} & DeepXplore~\cite{deepxplore} & DeepTest~\cite{tian2017deeptest} & DeepCover~\cite{SHK2018} & DeepGauge~\cite{deepgauge} \\ \hline
        Coverage criteria & NC, SSC, NBC etc.  & NC & NC & MC/DC & NBC etc. \\ \hline
        Test generation & concolic & dual-optimisation & greedy search & symbolic execution & gradient descent methods \\ \hline
        DNN inputs & single & multiple & single & single & single \\ \hline
        Image inputs & single/multiple & multiple & multiple & multiple & multiple \\ \hline
        Distance metric & $L_{\infty}$ and $L_0$-norm & $L_{1}$-norm & Jaccard distance & $L_{\infty}$-norm & $L_{\infty}$-norm \\
        %separation between oracle and testing objective & yes & no & yes & yes & yes \\ \hline
    \bottomrule
    \end{tabular}
    }
\end{table}

\subsection{Comparison with Related Work}

%Before proceeding to the experiments,
We briefly summarise the similarities and differences between our concolic
testing method, named \emph{DeepConcolic}, and other existing coverage-driven DNN testing methods: 
DeepXplore~\cite{deepxplore}, DeepTest~\cite{tian2017deeptest}, DeepCover~\cite{SHK2018},
and DeepGauge~\cite{deepgauge}. The details are presented in Table~\ref{tab:testings}, where NC, SSC, and NBC are short for Neuron Coverage, SS Coverage, and Neuron Boundary Coverage, respectively. In addition to the concolic nature of DeepConcolic, we observe the following differences.
\begin{itemize}
    \item DeepConcolic is generic, and is able to take coverage requirements as input; the other methods are \emph{ad hoc}, and are tailored to specific requirements.
    \item DeepXplore requires a set of DNNs to explore multiple gradient directions. The other methods, including DeepConcolic, need a single DNN only. 
    \item In contrast to the other methods, DeepConcolic can achieve good coverage by starting from a single input; the other methods need a non-trivial set of inputs.   
    \item Until now, there is no conclusion on the best distance metric. DeepConcolic 
    can be parameterized with a desired norm distance metric $\distance{\cdot}{}$. 
\end{itemize}
Moreover, DeepConcolic features a clean separation between the generation of test inputs and the test oracle. This is a good fit for traditional test case generation. The other methods use the oracle as part of their objectives to guide the generation of test inputs.

\section{Deep Neural Networks}

\newcommand{\real}{\mathds{R}}
\newcommand{\integer}{\mathds{N}}

A (feedforward and deep) neural network, or DNN, is a tuple $\networks=(L,
T, \Phi)$ such that 
%, where each of its elements is defined as follows.
$L=\{L_k|k\in\{1,\dots,K\}\}$ is a set of layers, 
$T\subseteq L\times L$ is a set of connections between layers, and
$\Phi=\{\phi_k|k\in\{2,\dots,K\}\}$ is a set of \emph{activation functions}.
Each layer $L_k$ consists of $s_k$ \emph{neurons}, and 
the $l$-th neuron of layer $k$ is denoted by $n_{k,l}$. 
We use $v_{k,l}$ to denote the value of $n_{k,l}$.
        Values of neurons in hidden layers (with $1<k<K$) need to pass through a 
        Rectified Linear Unit (ReLU)~\cite{relu}. For convenience, we explicitly
        denote the activation value before the ReLU as $u_{k,l}$ such that
        \begin{equation}
          \label{eq:relu}
           v_{k,l}=ReLU(u_{k,l})=
            \begin{cases}
              u_{k,l} &\mbox{  if } u_{k,l}\geq 0 \\
              0 & \mbox{  otherwise}
            \end{cases}
        \end{equation}
      ReLU is the most popular activation function for neural networks.

        Except for inputs, every neuron is connected to neurons in the preceding
        layer by pre-defined weights such that $\forall 1<k\leq K,\forall 1\leq l\leq s_k$,
        \begin{equation}
          \label{eq:sum}
          u_{k,l}=\sum_{1\leq h \leq s_{k-1}} \{w_{k-1, h, l}\cdot v_{k-1,h}\} + b_{k,l}
        \end{equation}
        where $w_{k-1,h,l}$ is the pre-trained weight for the connection between
        $n_{k-1,h}$ (i.e., the $h$-th neuron of layer $k-1$) and $n_{k,l}$
        (i.e., the $l$-th neuron of layer $k$), and $b_{k,l}$ is the 
        %so-called 
        \emph{bias}. % for neuron $n_{k,l}$.  

      Finally, for any input, the neural network assigns a \emph{label}, that is, the index
      of the neuron of the output layer that has the largest value,
%
      %\begin{equation}
      %  \label{eq:label}
      i.e.,  $\mathit{label}=\mathrm{argmax}_{1\leq l\leq s_K}\{v_{K,l}\}$.
      %\end{equation}

\commentout{
\begin{itemize}
  \item $L=\{L_k|k\in\{1,\dots,K\}\}$ is a set of layers such that
    $L_1$ is the \emph{input} layer, $L_{K}$ is the \emph{output} layer,
    and layers other than input and output layers are called \emph{hidden layers}.
    
    \begin{itemize}
      \item Each layer $L_k$ consists of $s_k$ 
      %neurons, which are
       % also called 
        \emph{neurons}.
      \item The $l$-th neuron of layer $k$ is
        denoted by $n_{k,l}$. 
    \end{itemize}

  \item $T\subseteq L\times L$ is a set of connections between layers such that, 
    except for the input and output layers, each layer has an incoming connection and an
    outgoing connection.
  \item $\Phi=\{\phi_k|k\in\{2,\dots,K\}\}$ is a set of \emph{activation functions}
    $\phi_k:D_{L_{k-1}}\rightarrow D_{L_k}$, one for each non-input layer. 

    \begin{itemize}
      \item We use $v_{k,l}$ to denote the value of $n_{k,l}$.
      \item Except for inputs, every neuron is connected to neurons in the preceding
        layer by pre-defined weights such that $\forall 1<k\leq K,\forall 1\leq l\leq s_k$
        \begin{equation}
          \label{eq:sum}
          v_{k,l}=\sum_{1\leq h \leq s_{k-1}} \{w_{k-1, h, l}\cdot v_{k-1,h}\} + \delta_{k,l}
        \end{equation}

        $w_{k-1,h,l}$ is the pre-trained weight for the connection between
        $n_{k-1,h}$ (i.e., the $h$-th neuron of layer $k-1$) and $n_{k,l}$
        (i.e., the $l$-th neuron of layer $k$) and $\delta_{k,l}$ the 
        %so-called 
        \emph{bias} for neuron $n_{k,l}$.  
        %We note that the
        %functions can express both fully-connected functions and
        %convolutional functions. We omit the mathematical forms of other layers such as 
        %maxpooling for space reasons.

      \item Values of neurons in hidden layers need to pass through a 
        Rectified Linear Unit (ReLU) \cite{relu}, such that \emph{the final activation 
        value} of each neuron of hidden layers is defined as
        \begin{equation}
          \label{eq:relu}
           v_{k,l}=ReLU(v_{k,l})=
            \begin{cases}
              v_{k,l} &\mbox{  if } v_{k,l}> 0 \\
              0 & \mbox{  otherwise}
            \end{cases}
        \end{equation}

      ReLU is by far the most popular and effective activation function for neural networks.

    \item Finally, for any input, the neural network assigns a \emph{label}, that is, the index
      of the neuron of the output layer having the largest value
      \begin{equation}
        \label{eq:label}
        \mathit{label}=\mathrm{argmax}_{1\leq l\leq s_K}\{v_{K,l}\}
      \end{equation}
      %Let $\labels$ be the set of labels. 
    \end{itemize}

\end{itemize}
}

\commentout{
\begin{example}
Figure \ref{fig:nn} is a simple neural network with four layers. 
Its input space is $D_{L_1}=\real^2$ where $\real$ the set of real numbers.

\begin{figure}[htp!]
\centering

\def\layersep{1.8cm}

\scalebox{0.7}{
\begin{tikzpicture}[shorten >=1pt,->,draw=black!50, node distance=\layersep]
    \tikzstyle{every pin edge}=[<-,shorten <=1pt]
    \tikzstyle{neuron}=[circle,fill=black!25,minimum size=15pt,inner sep=0pt]
    \tikzstyle{input neuron}=[neuron, fill=green!50];
    \tikzstyle{output neuron}=[neuron, fill=red!50];
    \tikzstyle{hidden neuron}=[neuron, fill=blue!50];
    \tikzstyle{annot} = [text width=4em, text centered]

    % Draw the input layer nodes
    \foreach \name / \y in {1,...,2}
    % This is the same as writing \foreach \name / \y in {1/1,2/2,3/3,4/4}
        \node[input neuron, pin=left:$v_{1,\y}$] (I-\name) at (0,-\y) {};
        %\node[input neuron, pin=left:Input \#\y] (I-\name) at (0,-\y) {};

    % Draw the 1st hidden layer nodes
    \foreach \name / \y in {1,...,3}
        \path[yshift=0.5cm]
            node[hidden neuron] (H1-\name) at (\layersep,-\y cm) {};

    % Draw the 2nd hidden layer nodes
    \foreach \name / \y in {1,...,3}
        \path[yshift=0.5cm]
            node[hidden neuron] (H2-\name) at (\layersep*2,-\y cm) {};

    % Draw the output layer node
    \node[output neuron,pin={[pin edge={->}]right:$v_{4,1}$}, right of=H2-2, yshift=0.5cm] (O1) {};
    \node[output neuron,pin={[pin edge={->}]right:$v_{4,2}$}, right of=H2-2, yshift=-0.5cm] (O2) {};

    % Connect every node in the input layer with every node in the
    % hidden layer.
    \foreach \source in {1,...,2}
        \foreach \dest in {1,...,3}
            \path (I-\source) edge (H1-\dest);

    \foreach \source in {1,...,3}
        \foreach \dest in {1,...,3}
            \path (H1-\source) edge (H2-\dest);

    \foreach \source in {1,...,3}
         \path (H2-\source) edge (O1);

    \foreach \source in {1,...,3}
         \path (H2-\source) edge (O2);

    % Annotate the layers
    \node[annot,above of=H1-1, node distance=1cm] (hl1) {Hidden layer};
    \node[annot,above of=H2-1, node distance=1cm] (hl2) {Hidden layer};
    \node[annot,left of=hl1] {Input layer};
    \node[annot,right of=hl2] {Output layer};

    \node[annot, right of=H1-1, node distance=0.0cm] (hl1) {\small $n_{2,1}$};
    \node[annot, right of=H1-2, node distance=0.0cm] (hl1) {\small $n_{2,2}$};
    \node[annot, right of=H1-3, node distance=0.0cm] (hl1) {\small $n_{2,3}$};
    \node[annot, right of=H2-1, node distance=0.0cm] (hl1) {\small $n_{3,1}$};
    \node[annot, right of=H2-2, node distance=0.0cm] (hl1) {\small $n_{3,2}$};
    \node[annot, right of=H2-3, node distance=0.0cm] (hl1) {\small $n_{3,3}$};
\end{tikzpicture}
}
  \caption{A simple neural network}
  \label{fig:nn}
\end{figure}

\end{example}
}

%Owing to the ReLU as in \eqref{eq:relu}, 
Due to the existence of ReLU, the neural network is a highly non-linear function.
% Given a neuron $n_{k,l}$, its ReLU is said to be
%\emph{activated} iff its value $v_{k,l}$ is positive; otherwise, when
%$ReLU(v_{k,l})=0$, the ReLU is deactivated.
%
%\paragraph{Neural network instance}
%
In this paper, we use variable $x$ to range over all possible inputs in the input domain
$D_{L_1}$ and use $t,t_1,t_2,...$ to denote concrete inputs.  
Given a particular input $t$, we say that the DNN $\networks$ is
instantiated and we use $\networks[t]$ to denote this instance of the network.

\begin{itemize}
  \item Given a network instance $\networks[t]$, the activation values of each neuron $n_{k,l}$ of the network 
    before and after ReLU are denoted as $u[t]_{k,l}$ and $v[t]_{k,l}$, respectively,
    and the final classification label is $label[t]$.
    We write $u[t]_k$ and $v[t]_k$ for $1\leq k\leq s_k$ to denote the vectors of activations for neurons in layer $k$. 
    \item When the input is given, the activation or deactivation of each ReLU operator in the DNN is determined. 

\commentout{
We may write $u[t]_{k,l}$ for the value before applying ReLU and $v[t]_{k,l}$ for the value after applying ReLU.
Moreover, we write  
  \begin{equation}
    \label{eq:sign}
    \mathit{sign}(v[t]_{k,l})=
    \begin{cases}
      +1 &\mbox{  if } u[t]_{k,l} = v[t]_{k,l} \\
      -1 & \mbox{  otherwise}
    \end{cases}
  \end{equation}
Intuitively, $\mathit{sign}(v[t]_{k,l})=-1$ means that the ReLU is not activated, and $\mathit{sign}(v[t]_{k,l})=+1$ otherwise. 
}
\end{itemize}

We remark that, while for simplicity the definition  focuses on DNNs with  
fully connected and convolutional layers,
as shown in the experiments (Section~\ref{sec:exp}) our method also 
applies to other popular layers,
e.g., maxpooling, used in state-of-the-art DNNs.

\newcommand{\boolf}{B}
\newcommand{\statf}{S}
\newcommand{\Var}{V}
\newcommand{\InputVar}{IV}
\newcommand{\subspace}{X}
\newcommand{\argopt}[4]{\arg #1_{#2}#3:  #4}

\section{Test Coverage for DNNs}
\label{sec:criteria}

\subsection{Activation Patterns}

A software program %problem
has a set of concrete execution paths.
Similarly, a DNN has a set of linear behaviours
%that are 
called \emph{activation patterns} \cite{SHK2018}.

\begin{definition}[Activation Pattern]
\label{def:activation-pattern}
Given a network $\networks$ and an input $t$, the activation pattern of
$\networks[t]$ is a function $ap[\networks,t]$ that maps the set of
hidden neurons to $\{\true,\false\}$. 
We write $ap[t]$ for $ap[\networks,t]$  if $\networks$ is clear from the context.
For an activation pattern $ap[t]$, we use $ap[t]_{k,i}$ to denote whether the ReLU operator of the neuron $n_{k,i}$ is activated or not. Formally,  
\begin{equation}\label{equ:ap}
\begin{array}{lcl}
ap[t]_{k,l} = \false & \equiv & u[t]_{k,l} < v[t]_{k,l}\\
ap[t]_{k,l} = \true  & \equiv &  u[t]_{k,l} =  v[t]_{k,l}
\end{array}
\end{equation} 
\end{definition}
%
%Given an input $t$, the $u[t]_{k,l}$ and $v[t]_{k,l}$ denote the activation value of $n_{k,l}$
%before and after the ReLU.
Intuitively, $ap[t]_{k,l} = \true$ if the ReLU  of the neuron $n_{k,l}$ is activated, and $ap[t]_{k,l} = \false$ otherwise.

Given a DNN instance $\networks[t]$, each ReLU operator's behaviour (i.e., each $ap[t]_{k,l}$)
is fixed and this results in the particular activation pattern $ap[t]$, which can be encoded 
by using a Linear Programming (LP) model \cite{SHK2018}.

Computing a test suite that covers all activation patterns of a DNN is intractable owing to the large number of neurons in pratically-relevant DNNs. Therefore,  we identify a subset of the activation patterns according to certain coverage criteria, and then generate test inputs that cover these activation patterns.

\subsection{Formalizing Test Coverage Criteria} 
\label{sec:qlpr}

We use a specific fragment of Quantified Linear Arithmetic over Rationals (QLAR) to express the coverage requirements on the test suite for a given DNN. This enables us to give a single test input generation algorithm (Section~\ref{sec:symbolic}) for a variety of coverage criteria. We denote the set of formulas in our fragment by \DRL.

\begin{definition}\label{def:requirement}
Given a network $\networks$, we write $\InputVar = \{x,x_1,x_2,...\} $ for a set of variables that range over the all inputs $D_{L_1}$ of the network.
We define $\Var = \{u[x]_{k,l},v[x]_{k,l}~|~ 1\leq
k\leq K, 1\leq l\leq s_k, x \in \InputVar\}$ to be a set of variables that range over the rationals.  We fix the following syntax for \DRL\ formulas:
\begin{equation}
\begin{array}{l}
r   ::=   ~ Q x. e~|~ Q x_1,x_2. e~ \\
%|~\argopt{opt}{x}{a}{e} ~|~\argopt{opt}{x_1,x_2}{a}{e}  \\
e   ::=   ~ a \bowtie 0~|~ e \land e ~|~ \neg e~|~ |\{e_1,...,e_m\}| \bowtie q\\
a   ::=   w ~|~ c \cdot w~|~p ~|~ a + a ~|~ a - a ~ \\
\end{array}
\end{equation}
where $Q\in \{\exists,\forall\}$, $w\in \Var$, $c,p\in \real$, $q\in\integer$, 
%, 
$\bowtie\in \{\leq,<, = , >,\geq\}$,  and $x,x_1,x_2\in \InputVar$. We call $r$ a coverage requirement, $e$ a Boolean formula, and $a$ an arithmetic formula. We call the logic \DRLp\ if the negation operator $\neg$ is not allowed. 
%\DRLe\ if the $opt$ operators are not allowed, and \DRLep\ if both $opt$ operators and the negation operator $\neg$ are not allowed.   
\emph{We use $\mathfrak{R}$ to denote a set of coverage requirement formulas.}  
\end{definition}

The formula $\exists x. e$ expresses that there exists an input
$x$ such that $e$ is true, while $\forall x. e$ expresses that $e$ is true for all inputs~$x$.  The
formulas $\exists x_1,x_2. e$ and  $\forall x_1,x_2. e$ have similar meaning, except that they quantify
over two inputs $x_1$ and $x_2$. The Boolean expression $|\{e_1,...,e_m\}|
\bowtie q$ is true if the number of true Boolean expressions in the set 
$\{e_1,...,e_m\}$ is in relation $\bowtie$ with $q$.  The other operators in
Boolean and arithmetic formulas have their standard meaning.

Although $V$ does not include variables to specify an activation pattern $ap[x]$, we may write
\begin{equation}
ap[x_1]_{k,l} = ap[x_2]_{k,l}\text{ and }ap[x_1]_{k,l} \neq ap[x_2]_{k,l}
\end{equation} 
to require that $x_1$ and $x_2$ have, respectively, the same and different activation
behaviours on neuron $n_{k,l}$.  These conditions can be expressed in the syntax above using the expressions in Equation~\eqref{equ:ap}.
Moreover, some norm-based distances between two inputs can be expressed using our syntax. For example, we can use the set of constraints
\begin{equation}
\label{eq:chebyshev}
\{x_1(i) - x_2(i) \leq q,\, x_2(i) - x_1(i) \leq q~|~ i\in \{1,\ldots,s_1\}\}
\end{equation} 
to express $\distance{x_1 - x_2}{\infty} \leq q$, i.e., we can constrain the Chebyshev distance
$L_\infty$ between two inputs $x_1$ and $x_2$, where $x(i)$ is the $i$-th
dimension of the input vector~$x$.
%
%, or structural similarity distances, such as SSIM \cite{WSB2003}. 
%The distance measure $L^p$ could be $L^1$ (Manhattan distance), $L^2$
%(Euclidean distance), $L^\infty$ (Chebyshev distance), etc.

%Depends on the requirements, we may consider either a Boolean test or a statistic test. 

%\begin{definition}\label{def:generalcriteria}
%
%A Boolean test is a function $\boolf: \mathfrak{R} \times D_{L_1}  \rightarrow \{0,1\}$, and a statistic test is a function $\statf: \mathfrak{R} \times D_{L_1}  \rightarrow \real$. 
% with respect to the criterion $C$. 
%\end{definition}
%
%Intuitively, $\boolf(r,x)$ indicates whether the requirement $r$ is completely satisfied by the input $x$, and $\statf(r,x)$ provides an evaluation on the satisfiability of $r$ by $x$.  

\newcommand{\xconstant}{t}

\subsubsection*{Semantics} 
\label{sec:semantics}

We define the satisfiability of a coverage requirement~$r$ by a test suite $\testsuites$.
%
%We assume that a variable will not be quantified more than once in a formula. For example, formula $\exists x. (e_1 \land \exists x. e_2) $ is not allowed, while $\exists x_1. (e_1 \land \exists x_2. e_2) $ is allowed. Those formulas in which variables are quantified more than once can be easily transformed into its legitimate  form by variable substitutions.  
%
%We use $\xconstant$ to represent a concrete input in a test suite $\testsuites$. 

\begin{definition}\label{def:testsemantics}
Given a set $\testsuites$ of test inputs and a coverage requirement $r$, the
satisfiability relation $\testsuites \models r$ is defined as follows.
\begin{itemize}
\item $\testsuites \models \exists x.e$ if there exists some test $\xconstant\in \testsuites$ such that $\testsuites \models e[x\mapsto \xconstant]$, where $e[x\mapsto \xconstant]$ denotes the expression $e$ in which the occurrences of $x$ are replaced by $\xconstant$.
\item $\testsuites \models \exists x_1,x_2.e$ if there exist two tests $\xconstant_1,\xconstant_2\in \testsuites$  such that $\testsuites \models e[x_1\mapsto \xconstant_1][x_2\mapsto \xconstant_2]$
\end{itemize}
The cases for $\forall$ formulas are similar. For the evaluation of Boolean expression $e$ over an input $\xconstant$, we have 
\begin{itemize}
\item $\testsuites \models a \bowtie 0$ if $a \bowtie 0$
\item $\testsuites \models e_1 \land e_2$ if $\testsuites \models e_1$ and $\testsuites \models e_2$
\item $\testsuites \models \neg e$ if not $\testsuites \models e$
\item $\testsuites \models |\{e_1,...,e_m\}| \bowtie q$ if $|\{e_i~|~\testsuites \models e_i, i \in \{1,...,m\}\}| \bowtie q$
\end{itemize} 
For the evaluation of arithmetic expression $a$ over an input $\xconstant$, % we have 
\begin{itemize}
\item $u[\xconstant]_{k,l}$ and $v[\xconstant]_{k,l}$ derive their values from the activation patters of the DNN for test $t$, and $c \cdot u[\xconstant]_{k,l}$ and $c \cdot v[\xconstant]_{k,l}$ have the standard meaning where $c$ is a coefficient, 
\item $p$, $a_1+a_2$, and $a_1-a_2$ have the standard semantics. 
\end{itemize} 

\end{definition}
Note that $\testsuites$ is finite. It is trivial to extend the definition of the satisfaction relation to an infinite subspace of inputs.

\subsubsection*{Complexity}

Given a network $\networks$, a \DRL\ requirement formula $r$, and a test suite $\testsuites$, checking $\testsuites\models r$ can be done in time that is polynomial in the size of $\testsuites$. Determining whether there exists a test suite $\testsuites$ with $\testsuites \models r$ is NP-complete.

\subsection{Test Coverage Metrics}

Now we can define test coverage criteria by providing a set of requirements on the test suite. The coverage metric is defined in the standard way as the percentage of the test requirements that are satisfied by the test cases in the test suite~$\testsuites$. 

\begin{definition}[Coverage Metric]
Given a network $\networks$, a set $\mathfrak{R}$ of test coverage requirements expressed as \DRL\ formulas, and a test suite $\testsuites$, the test coverage metric $\metric(\mathfrak{R},\testsuites)$ is as follows:
\begin{equation}
  \label{eq:madc}
  M(\mathfrak{R},\testsuites)=\frac{|\{ r \in \mathfrak{R} ~|~ \testsuites \models r \}|}{| \mathfrak{R}  |}
\end{equation}
\end{definition}

The coverage is used as a proxy metric for the confidence in the safety of the DNN under test.

\section{Specific Coverage Requirements} \label{sec:concreteRequirement}

In this section, we give \DRLp\ formulas for several important coverage criteria for
DNNs, including Lipschitz continuity~\cite{WB2015, ACB2017, WHK2018, BSZ2017, RHK2018} and test coverage criteria from the literature~\cite{deepxplore, SHK2018, deepgauge}. The criteria we consider have syntactical similarity with structural test coverage criteria in conventional software testing. Lipschitz continuity is semantic, specific to DNNs, and has been shown to be closely related to the theoretical understanding of convolutional DNNs~\cite{WB2015} and the robustness of both DNNs~\cite{WHK2018, RHK2018} and Generative Adversarial Networks~\cite{ACB2017}.  These criteria have been studied in the literature using a variety of formalisms and approaches.

Each test coverage criterion gives rise to a set of test coverage requirements. In the following, we discuss the three coverage criteria from~\cite{deepxplore, SHK2018, deepgauge}, respectively. 
We use $\distance{t_1-t_2}{q}$ to denote the distance between two inputs
$t_1$ and $t_2$ with respect to a given distance metric $\distance{\cdot}{q}$. 
The metric $\distance{\cdot}{q}$ can be, e.g., a norm-based metric such as
the $L_0$-norm (the Hamming distance), the $L_2$-norm (the Euclidean
distance), or the $L_\infty$-norm (the Chebyshev distance), or a
structural similarity distance, such as SSIM \cite{WSB2003}.  In the following,
we fix a distance metric and simply write $\distance{t_1-t_2}{}$. 
Section~\ref{sec:exp} elaborates on the particular metrics we use for our experiments.

We may consider requirements for a set of input subspaces.  Given a real
number $b$, we can generate a finite set $\mathcal{S}(D_{L_1},b)$ of
subspaces of $D_{L_1}$ such that for all inputs $x_1,x_2\in D_{L_1}$, if
$\distance{x_1-x_2}{} \leq b$, then there exists a subspace $\subspace \in
\mathcal{S}(D_{L_1},b)$ such that $x_1,x_2\in \subspace$.  The subspaces can
be overlapping.  Usually, every subspace $\subspace \in
\mathcal{S}(D_{L_1},b)$ can be represented with a box constraint, e.g.,
$\subspace = [l,u]^{s_1}$, and therefore $t \in \subspace$ can be expressed
with a Boolean expression as follows.
\begin{equation}
\bigwedge_{i=1}^{s_1} x(i) -u \leq 0 \land x(i) -l \geq 0
\end{equation} 

\subsection{Lipschitz Continuity}

In~\cite{SZSBEGF2014, RHK2018}, Lipschitz continuity has been shown to hold for a large class of DNNs, including DNNs for image classification.

\begin{definition}[Lipschitz Continuity]
	A network $\networks$ is said to be \emph{Lipschitz continuous} 
	if there exists a real constant $c\geq0$ such that, for all $x_1, x_2 \in D_{L_1}$:
	\begin{equation}
	 \distance{v[x_1]_1 -  v[x_2]_1}{} \le c \cdot \distance{x_1 -  x_2}{}
	\end{equation}
	Recall that $v[x]_1$ denotes the vector of activation values of the neurons in the input layer.
	The value $c$ is called the \emph{Lipschitz constant},
	and the smallest such $c$ is called the \emph{best Lipschitz constant}, denoted as $c_\mathit{best}$. 
\end{definition}

Since the computation of $c_\mathit{best}$ is an NP-hard problem and a smaller $c$
can significantly improve the performance of verification algorithms~\cite{WHK2018, WWRHK2018, RHK2018}, it is interesting to determine whether a given $c$ is
a Lipschitz constant, either for the entire input space $D_{L_1}$
or for some subspace.
Testing for Lipschitz continuity can be guided using the following requirements.

\begin{definition}[Lipschitz Coverage]
Given a real $c > 0$ and an integer $b > 0$, the set $\mathfrak{R}_{Lip}(b,c)$ of requirements for Lipschitz coverage is 
\begin{equation}
\label{eq:libr}
\begin{array}{l}
\{\exists x_1,x_2. (\distance{v[x_1]_1 - v[x_2]_1}{} - c \cdot \distance{x_1 - x_2}{} > 0)  \\
\hfill \land x_1,x_2 \in \subspace   ~|~ \subspace \in \mathcal{S}(D_{L_1},b)\}
\end{array}
\end{equation}
where the $\mathcal{S}(D_{L_1},b)$ are given input subspaces.
\end{definition} 
Intuitively, for each $\subspace \in \mathcal{S}(D_{L_1},b)$, this requirement expresses 
the existence of two inputs $x_1$ and $x_2$ that refute that $c$ is a Lipschitz constant for $\networks$.
%
%While  in theory we want to find a constant $c$ such that $M(\mathfrak{R}(b,c),\testsuites) = 0 $, in practice it can be sufficient to make $M(\mathfrak{R}(b,c),\testsuites) < \epsilon$ for some error bound $\epsilon$. 
%Then, the satisfiability of such a requirement means that the actual Lipschitz constant $c_{best}$ is larger than the given $c$. 
%
It is typically impossible to obtain full Lipschitz coverage, because there may exist inconsistent $r \in \mathfrak{R}_{Lip}(b,c)$. Thus, the goal for a test case generation algorithm is to produce a test suite~$\testsuites$ that satisfies the criterion as much as possible.

\subsection{Neuron Coverage}

Neuron Coverage (NC)~\cite{deepxplore} is an adaptation of statement coverage in conventional software testing to DNNs. It is defined as follows.

\begin{definition} %(Neuron Cover)
Neuron coverage for a DNN $\networks$ requires a test suite $\testsuites$ such that,
for any hidden neuron $n_{k,i}$, there exists test case $t\in \testsuites$ such that $ap[t]_{k,i} = \true$. 
\end{definition}

This is formalised with the following requirements $\mathfrak{R}_{NC}$, each of which expresses that there is a test with an input~$x$ that activates the neuron $n_{k,i}$, i.e., $ap[x]_{k,i} = \true$.

\begin{definition}[NC Requirements]
The set $\mathfrak{R}_{NC}$ of coverage requirements for Neuron Coverage is
\begin{equation}
\label{eq:ncrs}
\begin{array}{l}
%\{ \exists x. ~c_k\cdot ap[x]_{k,l}=\true ~|~  \hfill 2\leq k\leq K-1, 1\leq l\leq s_k\}
\{ \exists x.ap[x]_{k,i}=\true ~|~  \hfill 2\leq k\leq K-1, 1\leq i\leq s_k\}
\end{array}
\end{equation} 
\end{definition}

%Then the criteria is $M(\mathfrak{R}_{NC},\testsuites) > 1 - \epsilon$ for some error bound $\epsilon$. 

\subsection{Modified Condition/Decision (MC/DC) Coverage}

In~\cite{SHK2018}, a family of four test criteria is proposed, inspired by MC/DC coverage
in conventional software testing. We will restrict the discussion here to Sign-Sign Coverage (SSC). According to~\cite{SHK2018}, each neuron $n_{k+1,j}$ can be seen as a \emph{decision} where the neurons in the previous layer (i.e., the $k$-th layer) are conditions that define its activation value, as in Equation~\eqref{eq:sum}. Adapting MC/DC to DNNs, we must show that all condition neurons
can determine the outcome of the decision neuron independently. In the case of SSC coverage we say that the value of a decision or condition neuron changes if the sign of its activation function changes.
Consequently, the requirements for SSC coverage are defined as follows.
\begin{definition}[SSC Requirements]
For SCC coverage, we first define a requirement $\mathfrak{R}_{SSC}(\neuronpair)$ for a pair of neurons $\neuronpair=(n_{k,i},n_{k+1,j})$:
\begin{equation}
\label{eq:sscr}
\begin{array}{l}
\{ \exists x_1,x_2. ~ap[x_1]_{k,i} \neq ap[x_2]_{k,i} \land ap[x_1]_{k+1,j} \neq ap[x_2]_{k+1,j} \land \\
 \hfill \bigwedge_{1\leq l\leq s_k, l\neq i} ap[x_1]_{k,l} - ap[x_2]_{k,l} =0 \}
\end{array}
\end{equation}
and we get
\begin{equation}
  \label{eq:sscrs}
  \mathfrak{R}_{SSC} = \bigcup_{2\leq k\leq K-2, 1\leq i\leq s_k, 1\leq j\leq s_{k+1}}\mathfrak{R}_{SSC}((n_{k,i},n_{k+1,j}))
\end{equation}
\end{definition}

That is, for each pair $(n_{k,i},n_{k+1,j})$ of neurons in two adjacent layers $k$ and $k+1$, we need two inputs $x_1$ and $x_2$ such that the sign change of $n_{k,i}$ independently affects the sign change of $n_{k+1,j}$. Other neurons at layer $k$ are required to maintain their signs between $x_1$ and $x_2$ to ensure that the change is independent.  The idea of SS Coverage (and all other criteria in~\cite{SHK2018}) is to ensure that not only the existence of a feature needs to be tested but also the effects of less complex features on a more complex feature must be tested.

\subsection{Neuron Boundary Coverage}

Neuron Boundary Coverage (NBC)~\cite{deepgauge} aims at covering neuron activation values that exceed a given bound. It can be formulated as follows.

\begin{definition}[NBC Requirements]
%Given two sets of bounds $h=\{h_{k,i}\}_{2 \leq k\leq K-1,1\leq i\leq s_k}$ and $l=\{l_{k,i}\}_{2\leq k\leq K-1,1\leq i\leq s_k}$, the set $\mathfrak{R}_{NBC}(h,l)$ of requirements is
Given two sets of bounds $h=\{h_{k,i} | 2 \leq k\leq K-1,1\leq i\leq s_k \}$ and $l=\{l_{k,i} | {2\leq k\leq K-1,1\leq i\leq s_k} \}$, the requirements $\mathfrak{R}_\mathit{NBC}(h,l)$ are
\begin{equation}\label{eq:nbc}
    \begin{array}{l}
      \{ \exists x. ~u[x]_{k,i} - h_{k,i} > 0,\,\,\, \exists x.~u[x]_{k,i} - l_{k,i} < 0 ~|~ \\
       \hfill 2\leq k\leq K-1, 1\leq i\leq s_k \}
    \end{array}
\end{equation}
where $h_{k,i}$ and $l_{k,i}$ are the upper and lower bounds on the activation value of a neuron $n_{k,i}$.
\end{definition}
%
%As in \cite{deepgauge}, the $h_{k,i}$ and $l_{k,i}$ are estimated by from
%the training dataset.

%\subsection{Other Requirements} 

%$\mathfrak{T}$ the set of test suites.

\section{Overview of our Approach}
\label{sec:design}

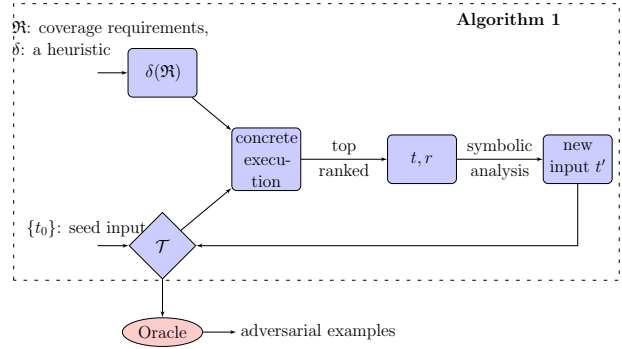
\begin{figure}[H]
    \centering
    \scalebox{.46}{\tikzstyle{decision} = [diamond, draw, fill=blue!20, 
    text width=4.5em, text badly centered, node distance=3cm, inner sep=0pt]
\tikzstyle{block} = [rectangle, draw, fill=blue!20, 
    text width=5em, text centered, rounded corners, minimum height=4em]
\tikzstyle{every node}=[font=\Large]
\tikzstyle{line} = [draw, -latex']
\tikzstyle{cloud} = [draw, ellipse,fill=red!20, node distance=3cm,
    minimum height=2em]

\begin{tikzpicture}[node distance = 2cm, auto]

  \node [] (init) {}; %$\{t_0\}$: the seed input};
  \node [above of=init, node distance=0.5cm, xshift=-0.2cm] (init2) {$\{t_0\}$: seed input};

  \node [decision, right of=init, node distance=2cm] (testsuite) {$\mathcal{T}$};
  
  \node [above of=init, node distance=5cm] (requirements) {};
  
  \node [above of=requirements, node distance=0.75cm, xshift=0.5cm, yshift=0.2cm] (requirements2) {
                                      \begin{tabular}{l}
                                        $\mathfrak{R}$: coverage requirements,\\$\delta$: a heuristic
                                      \end{tabular}
                                      };
  \node [block, right of=requirements, node distance=2cm] (requirements2) {$\delta(\mathfrak{R})$};
  
  \node [block, right of=testsuite, node distance=3cm, yshift=2.5cm] (concrete) {concrete execution};
  
  %\node [block, right of=concrete, node distance=4.5cm] (top-pair) {$(t,\delta(r))$};
  
  \node [block, right of=concrete, node distance=4.5cm] (top-pair) {$t, r$};
  
  \node [block, right of=top-pair, node distance=4.5cm] (testcase) {new input $t'$};

  \node [cloud, below of=testsuite, node distance=2.5cm] (oracle) {Oracle};
  
  \node [right of=oracle, node distance=4.5cm] (adversarials) {adversarial examples};
  
  \node [above of=top-pair, node distance=4cm, xshift=2.5cm] (algorithm) {\textbf{Algorithm 1}};

  % Draw edges
  \path [line, line width=0.3mm] (init) -- (testsuite);
  \path [line, line width=0.3mm] (requirements) -- (requirements2);
  \path [line, line width=0.3mm] (requirements2) -- (concrete);
  \path [line, line width=0.3mm] (testsuite) -- (concrete);
  \path [line, line width=0.3mm] (concrete) -- node[above]{top} node[below]{ranked} (top-pair);
  \path [line, line width=0.3mm] (top-pair) -- node[above]{symbolic} node[below]{analysis} (testcase);
  \path [line, line width=0.3mm] (testcase) |- (testsuite);
  \path [line, line width=0.3mm] (testsuite) -- (oracle);
  \path [line, line width=0.3mm] (oracle) --  (adversarials);
  
  \draw[loosely dashed] (-2.3,-1) rectangle (15.2,7);
  
\end{tikzpicture}}
    \caption{Overview of our concolic testing method}
    \label{fig:flow}
\end{figure}

This section gives an overview of our method for generating a test suite for a given DNN. Our method alternates between concrete evaluation of the activation patterns of the DNN and symbolic generation of new inputs. The pseudocode for our method is given as Algorithm~\ref{algo:testing}. It is visualised in Figure~\ref{fig:flow}. 

Algorithm~\ref{algo:testing} takes as inputs a DNN $\networks$, an input
$t_0$ for the DNN, a heuristic $\transform$, and a set $\mathfrak{R}$ of coverage requirements, and produces a test suite $\testsuites$ as output. The test suite $\testsuites$ initially only contains the given test input $t_0$. The algorithm removes a requirement $r\in\mathfrak{R}$ from $\mathfrak{R}$ once it is satisfied by $\testsuites$, i.e., $\testsuites\models r$.

\begin{algorithm}[!htp]
  \caption{Concolic Testing for DNNs}
  \label{algo:testing}
  \begin{flushleft}
    \textbf{INPUT:} $\networks, \mathfrak{R}, \transform, t_0$\\
    \textbf{OUTPUT:} $\testsuites$
  \end{flushleft}

  \begin{algorithmic}[1]
    \State $\testsuites\leftarrow\{t_0\}$ and $F=\{\}$
    \State $t\leftarrow t_0$
    \While{$\mathfrak{R} \setminus S \neq\emptyset$}
     \For{each $r\in\mathfrak{R}$}
       \If{$\testsuites\models r$} { $\mathfrak{R}\leftarrow\mathfrak{R}\setminus\{r\}$}
     \EndIf
     \EndFor
      \While{$\true$}
      %\State $t, \transform(r)\leftarrow requirement\_evaluation(\testsuites,\transform(\mathfrak{R}))$
      \State $t, r\leftarrow \mathit{requirement\_evaluation}(\testsuites,\transform(\mathfrak{R}))$
      \State $t' \leftarrow \mathit{symbolic\_analysis}(t,r) $
      %\State $t' \leftarrow symbolic\_analysis(t,\transform(r)) $
      %lp\_solve(\mathcal{C}[x][r], \min||x'-x_0||_{\infty})$
        \If{ $\mathit{validity\_check}(t') = \true$}
          \State $\testsuites\leftarrow\testsuites\cup\{t'\}$
          \State \textbf{break}
        \ElsIf{cost exceeded}
          \State $F\leftarrow F\cup\{r\}$
          \State \textbf{break}
        \EndIf
      \EndWhile
    \EndWhile
    \State \Return $\testsuites$
  \end{algorithmic}
  
\end{algorithm}

%we fail
%to mutate $t'$ so as to satisfy the requirement $r'$. 
%We define the mutation rules later.
%

The function $requirement\_evaluation$ (Line 7), whose details are given in
Section~\ref{sec:quantification}, looks for a pair
%$(t,\transform(r))$
$(t,r)$ \footnote{For some requirements, we might return two
inputs $t_1$ and $t_2$.  Here, for simplicity, we describe the case for a single
input.  The generalisation to two inputs is straightforward.} of input and
requirement that, according to our concrete evaluation, is the most
promising candidate for a new test case $t'$ that satisfies the requirement~$r$. 
The heuristic $\transform$ is a transformation function that maps a
formula $r$ with operator $\exists$ to an optimisation problem. 
This step relies on concrete execution. 

After obtaining $(t, r)$, 
$\mathit{symbolic\_analysis}$ (Line 8), whose details are in
Section~\ref{sec:symbolic}, is applied to obtain a new concrete input~$t'$. 
Then a function $\mathit{validity\_chec}k$ (Line 9), whose details are given in
Section~\ref{sec:oracle}, is applied to check whether the new input is valid or
not. If so, the test is added to the test suite. Otherwise, ranking and symbolic input generation are repeated until a given computational cost is exceeded, after which test generation for the requirement is deemed to have failed. This is recorded in the set~$F$.

% that satisfies a transformed requirement $\transform(r)$.  
%
%Then, we choose a pair of input-requirement $(t,r)$ by considering all test cases in
%$\testsuites$ and all requirements in $\mathfrak{R}$ such that the evaluation result
%$eval(t,r)$ is the largest. Intuitively, among all requirements that have not been
%satisfied, $r$ is regarded as the one that is most likely to be satisfied in the next,
%by mutating $x$.
%
%Given $x$ and $r$, $\mathcal{C}[x][r]$ is the mutated set of constraints of $\mathcal{C}[x]$
%such that $r$ is satisfied now. If the mutated constraint set is satisfied, then a new input
%$x'$ is returned by the LP solver and it is added into the test suite $\testsuites$. Otherwise,
%$(x,r)$ becomes part of the unreachable pairs in $S$. When generating a new input, we try to keep
%it as close as possible to the original input $x_0$.
%
The algorithm terminates when either all test requirements have been satisfied, i.e.,
$\mathfrak{R}=\emptyset$, or no further requirement in $\mathfrak{R}$ can be
satisfied, i.e., $F =\mathfrak{R}$. It then returns the current test suite $\testsuites$.

 %Alternatively, if , the testing also stops
%and $\testsuites$ is returned. In the latter case, to improve the coverage of test requirements,
%the Algorithm  \ref{algo:testing} can be called with different input $x_0$.

Finally, as illustrated in Figure~\ref{fig:flow}, the test suite $\testsuites$ 
generated by Algorithm~\ref{algo:testing}, is passed to an oracle in order to evaluate the robustness
of the DNN. The details of the oracle are in Section~\ref{sec:oracle}.

\section{Ranking Coverage Requirements}
\label{sec:quantification}

This section presents our approach for Line 7 of
Algorithm~\ref{algo:testing}.  Given a set of requirements
$\mathfrak{R}$ that have not yet been satisfied, a heuristic~$\transform$, and the current set
$\testsuites$ of test inputs, the goal is to select a concrete input $t \in
\testsuites$ together with a requirement  $r\in
\mathfrak{R}$, both of which will be used later in a symbolic approach to
compute the next concrete input $t'$ (to be given in
Section~\ref{sec:symbolic}). The selection of $t$ and $r$ is done by means of a series of concrete executions.

The general idea is as follows. 
For all requirements $r\in\mathfrak{R}$, we transform $r$ into $\transform(r)$ by utilising operators $\arg
opt$ for $opt\in \{\max,\min\}$ that will be evaluated by concretely executing tests in~$\testsuites$.
%Then for all pairs $(t,\transform(r)) \in \testsuites\times \transform(\mathfrak{R})$, we apply the semantics in Definition~\ref{def:testsemantics} to obtain 
%an evaluation $val(t,\transform(r))$ (details to be given below). 
%
%Then obtain an input $t_r$ by applying
%semantics in Definition \ref{def:testsemantics} over $\transform(r)$ and
%$\testsuites$.  When returning $t_r$, an evaluation $val(t_r,r)$ will be
%returned as well.
%
%, which can then be used on the existing 
%test suite $\testsuites$ to obtain $x'$. The new requirement $r'$ may use operator $\arg opt$ for $opt \in \{\max,\min\}$, 
%whose evaluation on $\testsuites$ with respect to the semantics in Definition \ref{def:testsemantics} will generate the best available test case $x'$ in $\testsuites$.  
%
As $\mathfrak{R}$ may contain more than one requirement, we return the pair
$(t,r)$ such that
\begin{equation}
r = \arg\max_{r} \{val(t,\transform(r))~|~r\in \mathfrak{R}\}. 
\end{equation}
Note that, when evaluating $\arg opt$ formulas (e.g., $\arg\min_x a:e$), if an input $t\in\testsuites$ is returned, we may need the value ($\min_x a:e$) as well. We use $val(t,\transform(r))$ to denote such a value for the returned input $t$ and the requirement formula $r$. 

%\xiaowei{can be useful to attach with this paragraph a pseudo-code}
%are going to define an evaluation
%function, in order to quantitively estimate to what extent the requirement is covered.

The formula $\transform(r)$ is an optimisation objective together with a set of constraints. We will give several examples later in Section~\ref{sec:heuristics}. In the following, we extend the semantics in Definition~\ref{def:testsemantics} to work with formulas with $\arg opt$ operators for $opt\in\{\max,\min\}$, including $\arg opt_x a:e$ and $\arg opt_{x_1,x_2} a:e$. 
Intuitively, $\arg\max_{x} a: e$ ($\arg\min_{x} a: e$,
resp.) determines the input $x$ among those satisfying the Boolean formula $e$
that maximises (minimises) the value of the arithmetic formula~$a$. Formally, 

\begin{itemize}
\item the evaluation of $\arg\min_x a:e $ on $\testsuites$ returns an input $\xconstant\in \testsuites$ such that,  $\testsuites \models e[x\mapsto \xconstant]$ and for all $\xconstant'\in \testsuites$ such that $\testsuites \models e[x\mapsto \xconstant']$ we have $a[x\mapsto \xconstant] \leq  a[x\mapsto \xconstant']$. 
\item the evaluation of $\testsuites \models \arg\min_{x_1,x_2} a:e $ on $\testsuites$ returns two inputs $\xconstant_1,\xconstant_1\in \testsuites$ such that,  $\testsuites \models e[x_1\mapsto \xconstant_1][x_2\mapsto \xconstant_2]$ and for all $\xconstant_1',\xconstant_2'\in \testsuites$ such that $\testsuites \models e[x_1\mapsto \xconstant_1'][x_2\mapsto \xconstant_2']$ we have $a[x_1\mapsto \xconstant_1][x_2\mapsto \xconstant_2] \leq  a[x_1\mapsto \xconstant_1'][x_2\mapsto \xconstant_2']$. 
\end{itemize}
The cases for $\arg\max$ formulas are similar to those for $\arg\min$, by replacing $\leq$ with $\geq$. 
Similarly to Definition~\ref{def:testsemantics}, the semantics is for a set $\testsuites$ of test cases and we can  adapt it to a continuous input subspace
$X\subseteq D_{L_1}$.

%When evaluating a criterion, we do not consider
%formulas with operators $\arg opt$, which are used primarily in
%Section~\ref{sec:quantification} to express heuristics.

\subsection{Heuristics}\label{sec:heuristics}

We present the heuristics $\transform$ we use the coverage requirements discussed in Section~\ref{sec:concreteRequirement}.
%
%We require $r$ to be in
%a fragment of \DRLep\ in which the formulas can only have either $\geq$ and $>$, or $\leq$ and $<$. The additional restriction enables our automated translation of the problem into either a maximisation or a minimisation problem. 
We remark that, since $\transform$ is a heuristic, there exist alternatives. The following definitions work well in our experiments.  

%Now, for a Boolean expression $e$, we write $opt(e) = \max$, if $e$ includes only $\geq$ and $>$, and $opt(e)=\min$, if $e$ includes only $\leq$ and $<$. As any Boolean expression $e$ in \DRLep\ is a conjunctive formula, we let $cond(e)$ be the subexpression of $e$ which uses symbol $=$, and thus $e\setminus cond(e)$ be the remaining subexpression with symbols $\{\leq,<,>,\geq\}$.    

%The transformation $\transform(r)$ is done recursively over the structure of the formula $r$. 
% \xiaowei{but the following transformations are ad-hoc ...}
%
%\begin{equation}
%\begin{array}{l}
%\displaystyle\transform(\exists x.e) = \arg opt(e)_{x}. \transform(e\setminus cond(e)): cond(e) \\
%\transform(a \bowtie 0) = a \\
%\transform(|\{a_1 \bowtie 0,...,a_m\bowtie 0\}| \bowtie q)  =  \sum_{i=1}^m a_i\\

%\end{array}
%\end{equation}

%As examples, we provide transformations for several cases we discussed in Section~\ref{sec:concreteRequirement}. Intuitions are provided to shed light on the transformations. Other cases can be directly obtained by the above transformation. 

\subsubsection{Lipschitz Continuity}

When a Lipschitz requirement $r$ as in Equation (\ref{eq:libr}) is
not satisfied by $\testsuites$, we transform it into $\transform(r)$ as
follows:
\begin{equation}
\label{eq:libropt}
\arg\max_{x_1,x_2}. \distance{v[x_1]_1 - v[x_2]_1}{} - c * \distance{x_1 - x_2}{}: x_1,x_2 \in \subspace
\end{equation}
I.e., the aim is to find the best $t_1$ and $t_2$ in
$\testsuites$ to make $\distance{v[t_1]_1 - v[t_2]_1}{} - c \cdot
\distance{t_1 - t_2}{}$ as large as possible.  As described, we also need to
compute $val(t_1,t_2,r) = \distance{v[t_1]_1 - v[t_2]_1}{} - c \cdot \distance{t_1 -
t_2}{}$.

\commentout{

\subsubsection{Safety}

When a safety requirement $r$ in Definition~\ref{equ:boolsafety} is
unsatisfiable on $\testsuites$, we may transform it into the following
requirement $\transform(r)$:
%
%replacing Equation (\ref{equ:boolsafety}) with the following one:  
\begin{equation} 
\begin{array}{l}
\displaystyle \arg\max_{x} \sum_{i \in \{1, ..., s_k\}}(v[x]_{K,i} - v[x]_{K,l(\subspace)}): x \in \subspace 
\end{array}\label{equ:statisicafety}
\end{equation}
Intuitively, the aim is to find an input $t$ in $\testsuites$ which falls within
the region $X$ and is close to breaking the safety.  The computation of the value $val(t,r)$ is
straightforward, as for Lipschitz continuity.

}

\subsubsection{Neuron Cover}
%Given an individual requirement $\mathfrak{R}(n_{k,i})\in\mathfrak{R}_{NC}$, let us say that it requests
%the neuron $n_{k,i}$ be activated, that is, $r_{k,i}$ is $\exists x.ap[x]_{k,i}=1$. 
%Subsequently, given an input $x$, we define the following function
%to quantify to what level $r_{k,i}$ is satisfied.
%
When a requirement $r$ as in Equation~(\ref{eq:ncrs}) is not satisfied by
$\testsuites$, we transform it into the following requirement
$\transform(r)$:
\begin{equation}
  \label{eq:eval-nc}
  \argopt{\max}{x}{c_k \cdot u_{k,i}[x]}{\true}
\end{equation}
%
%\begin{equation}
%  \label{eq:eval-nc}
%  eval(x, \mathfrak{R}(n_{k,i}))=u_{k,i}[x]\cdot f_k
%\end{equation}
%
We obtain the input $t\in \testsuites$ that
has the maximal value for $c_k \cdot u_{k,i}[x]$. 

The coefficient $c_k$ is a per-layer constant. It motivated by the following observation.
With the propagation of signals in the DNN, activation values at each layer can
be of different magnitudes. For example, if the minimum activation value of neurons
at layer $k$ and $k+1$ are $-10$ and $-100$, respectively, then even when a neuron
$u[x]_{k,i}=-1>-2 = u[x]_{k+1,j}$, we may still regard $n_{k+1,j}$ as being closer to
be activated than $u_{k,i}$ is.
Consequently, we define a layer factor $c_k$ for each layer that
normalises the average activation valuations of neurons at different layers into the same
magnitude level. It is estimated by sampling a sufficiently large input dataset.

%The coefficient $c_k$ is
%a layer-wise constant.  It is based on the following observation.  With the
%propagation of signals in the DNN, activation values at each layer can be of
%different magnitudes.  For example, if the minimum activation value of
%neurons at layer $k$ and $k+1$ are -10 and -100 respectively, then even when
%a neuron $u[x]_{k,i}=-1>-2 = u[x]_{k+1,j}$, we may still regard $n_{k+1,j}$
%as being closer towards being activated than $u_{k,i}$ does.
%
%Consequently, we define a layer factor $c_k$ for each layer which normalises
%the average activation valuations of neurons at different layers into the
%same magnitude level.
%
%  the activation value of the
%assessed neuron $n_{k,i}$. Intuitively, if $eval$ returns the positive result, the NC
%requirement $r_{k,i}$ for the neuron $n_{k,i}$ is satisfied; otherwise, a larger value indicates
%$n_{k,i}$ is closer to be activated.

%\commentout{

\subsubsection{SS Coverage}

In SS Coverage, given a decision neuron $n_{k+1,j}$, the concrete evaluation aims to select one of 
its condition neurons $n_{k,i}$ at layer $k$ such that the test input that is generated negates the signs of $n_{k,i}$ and
$n_{k+1,j}$ while the remainder of $n_{k+1,j}$'s condition neurons preserve their respective
signs. This is achieved by the following $\transform(r)$:
\begin{equation}
  \label{eq:eval-ssc}
  \argopt{\max}{x}{- {c_k\cdot|u[x]_{k,i}|} }  {\true}
\end{equation}
Intuitively, given the decision neuron $n_{k+1,j}$, Equation (\ref{eq:eval-ssc}) selects the condition
that is closest to the change of activation sign  %change 
(i.e., yields the smallest $|u[x]_{k,i}|$).

%\youcheng{give up SSC of the whole DNN and instead focus on one particular decision neuron here}

\commentout{
\subsection{DS Cover}
As formulated in Equation (\ref{eq:dscr}), each requirement $\mathfrak{R}_{DS}(\alpha, h, q)\in\mathfrak{R}_{DS}$
specifies the sign change of one neuron $n_{k+1,j}$ at layer $k+1$ and the distance change of neurons
at layer $k$ by $h$ and $q$. We thus define $eval(x, \mathfrak{R}_{DS}(\alpha, h, q))$ as follows
\begin{equation}
  \label{eq:eval-dsc}
  eval(x, \mathfrak{R}_{DS}(\alpha, h, q))= \frac{\min_{1\leq i\leq s_k}\{|u[x]_{k,i}|\}\cdot c_k}{|u[x]_{k+1,j}|\cdot c_{k+1}}
\end{equation}
}

\subsubsection{Neuron Boundary Coverage}

We transform the requirement~$r$ in Equation (\ref{eq:eval-nbc}) into the following $\transform(r)$ when it is not satisfied by~$\testsuites$;
it selects the neuron that is closest to either the higher or lower boundary.
\begin{equation}
  \label{eq:eval-nbc}
  \begin{array}{l}
  \argopt{\max}{x}{ c_k\cdot (u[x]_{k,i}- h_{k,i})  } {\true} \\ \argopt{\max}{x}{  c_k\cdot(l_{k,i}-u[x]_{k,i})  } {\true}
  \end{array}
\end{equation}

%\begin{equation}
%  \label{eq:eval-ncm}
%  eval(x, \mathfrak{R}_{NC}^m(n_{k,i}))= \frac{|\{n_{k,j}| u[x]_{k,j}<u_[x]_{k,i}\}|}{s_k}
%\end{equation}

%}

\newcommand\scalemath[2]{\scalebox{#1}{\mbox{\ensuremath{\displaystyle #2}}}}

\section{Symbolic Generation of New Concrete Inputs}
\label{sec:symbolic}

This section presents our approach for Line~8 of Algorithm~\ref{algo:testing}. That is, given a concrete input $t$ and a  requirement $r$, we need to find the next concrete input $t'$ by symbolic analysis. This new $t'$ will be added into the test suite (Line~10 of Algorithm~\ref{algo:testing}). 
The symbolic analysis techniques to be considered include the linear programming in~\cite{SHK2018}, global optimisation for the $L_0$ norm in~\cite{RWSHKK2018}, and a new optimisation algorithm that will be introduced below.
%
%Because the transformed requirements in Section~\ref{sec:quantification} are optimisation problems, symbolic analysis will be designed to solve optimisation problems
We regard optimisation algorithms as symbolic analysis methods because, similarly to constraint solving methods, they work with a set of test cases in a single run. 

To simplify the presentation, the following description may, for each algorithm, focus on some specific coverage requirements, but we remark that all algorithms can work with all the requirements given in Section~\ref{sec:concreteRequirement}. 

%\xiaowei{Here we explain the approaches of using optimisation algorithms after having a concrete inuput}

\subsection{Symbolic Analysis using Linear Programming}

As explained in Section~\ref{sec:criteria}, given an input $x$,
the DNN instance $\networks[x]$ maps to an activation pattern $ap[x]$ that can be
modeled using Linear Programming (LP). In particular, the following linear constraints~\cite{SHK2018}
yield a set of inputs that exhibit the same ReLU behaviour as~$x$:
\begin{equation} \label{eq:lp-v}
\scalemath{0.9}{
\{  \mathbf{u_{k,i}}=\sum_{1\leq j \leq s_{k-1}} \{{w}_{k-1, j, i}\cdot \mathbf{v_{k-1,j}}\} + b_{k,i}~|~k\in [2,K], i\in [1..s_k]\}
}
\end{equation}
\begin{equation}
\label{eq:lp-dir1}
\scalemath{0.9}{
\begin{array}{l}
 \{\mathbf{u_{k,i}}\geq 0 \wedge \mathbf{u_{k,i}}=\mathbf{v_{k,i}} ~|~ ap[x]_{k,i}=\true, k\in [2,K), i\in [1..s_k] \}\\
\cup 
  \{\mathbf{u_{k,i}}< 0 \wedge \mathbf{v_{k,i}}=0~|~ap[x]_{k,i}=\false, k\in [2,K), i\in [1..s_k]\}
\end{array}
}
\end{equation}
Continuous variables in the LP model are emphasized in \textbf{bold}.
\begin{itemize}
    
    \item The activation value of each neuron is encoded by the linear constraint in~\eqref{eq:lp-v}, which
    is a symbolic version of Equation~\eqref{eq:sum} that calculates a neuron's activation value.
    
    \item Given a particular input $x$, the activation pattern (Definition \ref{def:activation-pattern}) $ap[x]$
    is known: $ap[x]_{k,i}$ is either $\true$ or $\false$, which indicates whether the ReLU is activated
    or not for the neuron~$n_{k,i}$.  Following~\eqref{equ:ap} and 
    the definition of ReLU in \eqref{eq:relu}, for every neuron $n_{k,i}$, the linear constraints in \eqref{eq:lp-dir1}
    encode ReLU activation (when $ap[x]_{k,i}=\true$) or deactivation (when $ap[x]_{k,i}=\false$).
\end{itemize}

The linear model (denoted as $\mathcal{C}$)
given by \eqref{eq:lp-v} and \eqref{eq:lp-dir1} represents an input set that
%come up with
results in the same activation pattern as encoded. Consequently,
the symbolic analysis for finding a new input $t'$ from a pair $(t, r)$ of 
input and requirement is equivalent to finding a new activation pattern. \emph{Note that,
to make sure that the obtained test case is meaningful, an objective
is added to the LP model that minimizes the distance between $t$ and $t'$.} Thus, the use of LP requires that the distance metric is linear. For instance, this applies to the $L_{\infty}$-norm in \eqref{eq:chebyshev}, but not to the $L_2$-norm.

\subsubsection{Neuron Coverage}
The symbolic analysis of neuron coverage takes the input test case $t$ and requirement $r$ on the activation of neuron $n_{k,i}$, and returns a new test $t'$ such that the test requirement
is satisfied by the network instance $\networks[t']$. We have the activation pattern
$ap[t]$ of the given $\networks[t]$, and can build up a new activation pattern $ap'$ such that 
\begin{equation}
\label{eq:lp-nc}
\{ ap'_{k,i}=\neg ap[t]_{k,i} \wedge \forall k_1<k:\bigwedge\limits_{0\leq i_1\leq s_{k_1}}\,\, ap'_{k_1,i_1}=ap[t]_{k_1,i_1}\}
\end{equation}
This activation pattern %by Equation \eqref{eq:lp-nc} 
specifies the following conditions.

\begin{itemize}
  \item 
  %The neuron 
  $n_{k,i}$'s activation sign is negated: this encodes the goal to activate $n_{k,i}$.
  \item In the new activation pattern $ap'$, the neurons before layer $k$ preserve their activation signs as in $ap[t]$. 
  Though there may exist multiple activation patterns that make $n_{k,i}$ activated, for the use of LP modeling one particular combination of activation signs must be pre-determined.
  \item Other neurons are irrelevant, as the sign of $n_{k,i}$ is only affected by the activation values of those neurons in previous layers.
  %the propagation
  %of signals through neurons in prior to it.
\end{itemize}

Finally, the new activation pattern $ap'$ defined in \eqref{eq:lp-nc} is encoded by the LP model $\mathcal{C}$ using \eqref{eq:lp-v} and \eqref{eq:lp-dir1}, and if there exists a feasible solution, then the new test input $t'$, which satisfies the requirement~$r$, can be extracted from that solution.

\subsubsection{SS Coverage}

To satisfy an SS Coverage requirement $r$, we need to find a new 
test case such that, with respect to the input~$t$, the activation signs of $n_{k+1,j}$ and $n_{k,i}$
are negated, while other signs of other neurons at layer $k$ are equal to those for input~$t$.

To achieve this, the following activation pattern $ap'$ is constructed.
\begin{equation*}
\begin{array}{l}
\label{eq:lp-ssc}
\{ ap'_{k,i}=\neg ap[t]_{k,i}  \land ap'_{k+1,j}=\neg ap[t]_{k+1,j}  \\ 
    \hfill \land \forall k_1<k:\bigwedge\limits_{1\leq i_1\leq s_{k_1}}\,\, ap'_{k_1,i_1}=ap[t]_{k_1,i_1}\}
\end{array}
\end{equation*}

%The new activation pattern negates the signs of the neuron pair $n_{k+1,j}$ and $n_{k,i}$, while it reserves the signs
%of other precedent neurons of $n_{k+1,j}$. If a solution exists in the LP model of $ap_2$, it will be the new input returned
%by the symbolic analysis.

\subsubsection{Neuron Boundary Coverage}

In case of the neuron boundary coverage, the symbolic analysis aims to find an input $t'$ 
such that the activation value of neuron $n_{k,i}$ exceeds either its higher bound $h_{k,i}$ or its
lower bound $l_{k,i}$.

To achieve this, while preserving the DNN activation pattern $ap[t]$, we
add one of the following constraints to the LP program.
\begin{itemize}
\item If $u[x]_{k,i}-h_{k,i}>l_{k,i}-u[x]_{k,i}$: ${u_{k,i}}>h_{k,i}$;
\item otherwise: ${u_{k,i}}<l_{k,i}$.
\end{itemize}

\subsection{Symbolic Analysis using Global Optimisation}
\label{sec:symbol-optimisation}

The symbolic analysis for finding a new input can also be implemented by solving the global optimisation
problem in~\cite{RWSHKK2018}. That is, by specifying the test requirement as an optimisation objective, we
apply global optimisation to compute a test case that satisfies the test coverage requirement.

%\youcheng{TBD: a brief description on the global optimisation}

\begin{itemize}
\item For Neuron Coverage, the objective is to find a $t'$ such that the specified neuron $n_{k,i}$
has ${ap[t']_{k,i}}=$\true.

\item In case of SS Coverage, given the neuron pair $(n_{k,i}, n_{k+1,j})$ and the original input $t$,
the optimisation objective becomes
\begin{equation*}
\begin{array}{l}
ap[t']_{k,i}\neq ap[t]_{k,i} \land ap[t']_{k+1,j}\neq \\\hfill ap[t]_{k+1,j}
\land \bigwedge\limits_{i'\neq i} ap[t']_{k,i'}= ap[t]_{k,i}
\end{array}
\end{equation*}
\item Regarding
%When it comes to 
the Neuron Boundary Coverage, depending on whether the higher bound or lower bound
for the activation of $n_{k,i}$ is considered, the objective of finding a new input $t'$ is either
$u[t']_{k,i}>h_{k,i}$ or $u[t']_{k,i}<l_{k,i}$.

\end{itemize}
Readers are referred to~\cite{RWSHKK2018} for the details of the algorithm. 

\subsection{Lipschitz Test Case Generation}

%\wenjie{This section is reserved for wenjie writing lipschitz test case generation, will delete this mark once finished}

Given a coverage requirement as in Equation (\ref{eq:libr}) for a subspace~$X$, we let $t_0 \in \mathbb{R}^n$ be the representative point of the subspace $X$ to which $t_1$ and $t_2$ belong. 
The optimisation problem is to generate two inputs $t_1$ and $t_2$ such that
\begin{equation}
\begin{split}
|| v[t_1]_{1} - v[t_2]_1||_{D_1}- c \cdot ||t_1 - t_2||_{D_1} > 0 \\
\mbox{s.t.~~} || t_1 - t_0 ||_{D_2} \leq \Delta , ~ || t_2 - t_0 ||_{D_2} \leq \Delta 
\end{split}
\end{equation}
where 
%xiaowei: I don't think software engineering people can understand this difference
%$f(t)$ represents the chosen logit $v[t]_{K-1}$ or probabilistic confidence output $v[t]_K$ of a network, 
$||*||_{D_1}$ and $||*||_{D_2}$ denote norm metrics such as the $L_0$-norm, $L_2$-norm or $L_{\infty}$-norm, and $\Delta$ is the radius of a norm ball (for the $L_1$ and $L_2$-norm) or the size of a hypercube (for the $L_{\infty}$-norm) centered on $t_0$. The constant $\Delta$ is a hyper-parameter of the algorithm. 

%We formulate the above satisfaction problem as an optimisation problem which can be 
The above problem can be efficiently solved by a novel {\em alternating compass search} scheme. Specifically, we alternate between solving the following two optimisation problems through relaxation~\cite{Roubicek1997}, i.e., maximizing the lower bound of the original Lipschitz constant instead of directly maximizing the Lipschitz constant itself. To do so, we reformulate the original non-linear proportional optimisation as a linear problem when both norm metrics $||*||_{D_1}$ and $||*||_{D_2}$ are the $L_\infty$-norm.
\subsubsection{Stage One} We solve
	\begin{equation}\label{eqn-27}
	\begin{split}
	\min_{t_1} F(t_1,t_0) =  - || v[t_1]_1 - v[t_0]_1||_{D_1} \\
	\text{s.t.~~} || t_1 - t_0 ||_{D_2} \leq \Delta
	\end{split}
	\end{equation}
	 The objective above enables the algorithm to search for an optimal $t_1$ in the space of a norm ball or hypercube centered on $t_0$ with radius $\Delta$, maximising the norm distance of $v[t_1]_1$ and $v[t_0]_1$. The constraint implies that $\sup_{ || t_1 - t_0 ||_{D_2} \leq \Delta}|| t_1 - t_0 ||_{D_2} = \Delta$. Thus, a smaller $F(t_1,t_0)$ yields a larger Lipschitz constant, considering that $\mathbf{Lip}(t_1,t_0) = -F(t_1,t_0)/ || t_1 - t_0 ||_{D_2} \geq -F(t_1,t_0)/ \Delta$, i.e., $ -F(t_1,t_0)/ \Delta$ is the lower bound of $\mathbf{Lip}(t_1,t_0)$. Therefore, the search for a trace that minimises $F(t_1,t_0)$ increases the Lipschitz constant.
	 
	To solve the problem above we use the \emph{compass search method}~\cite{audet2017derivative}, which is efficient, derivative-free, and
	guaranteed to provide first-order global convergence. Because we aim to find an input pair that refutes the given Lipschitz constant $c$ instead of finding the largest possible Lipschitz constant, %as a result, 
	along each iteration, when we get $\bar{t}_1$, we check 
	%satisfaction by evaluating 
	whether $ \mathbf{Lip}(\bar{t}_1,t_0) > c$. % holds. 
	If it holds, we find an input pair $\bar{t}_1$ and $t_0$ that satisfies the test requirement; otherwise, we continue the compass search until convergence  or a satisfiable input pair is generated. If Equation (\ref{eqn-27}) is convergent and we can find an optimal $t_1$ as
	$$
	t_1^* = \arg \min_{t_1} F(t_1,t_0) \text{~~~s.t.~} || t_1 - t_0 ||_{D_2} \leq \Delta
	$$
	but we still cannot find a satisfiable input pair, we perform the Stage Two optimisation.
	
	\subsubsection{Stage Two} We solve 
	\begin{equation}\label{eqn-28}
	\begin{split}
	\min_{t_2} F(t_1^*,t_2) =  - || v[t_2]_1 - v[t_1^*]_1||_{D_1} \\
	\text{s.t.~~} || t_2 - t_0 ||_{D_2} \leq \Delta
	\end{split}
	\end{equation}
	Similarly, we use derivative-free compass search to solve the above problem and check whether $\mathbf{Lip}(t_1^*,t_2)> c$ holds at each iterative optimisation trace $\bar{t}_2$. If it holds, we return the image pair $t_1^*$ and $\bar{t}_2$ that satisfies the test requirement; otherwise, we continue the optimisation until convergence or a satisfiable input pair is generated. If Equation (\ref{eqn-28}) is convergent at $t_2^*$, and we still cannot find such a input pair, we modify the objective function again by letting $t_1^* = t_2^*$ in Equation (\ref{eqn-28}) and continue the search and satisfiability checking procedure.
	
	\subsubsection{Stage Three} 
	%In Stage Two, no matter how we alternatingly change the objective function,
	If the function $\mathbf{Lip}(t_1^*,t_2^*)$ fails to make progress in Stage Two, we treat the whole search procedure as convergent and have failed to find an input pair that can refute the given Lipschitz constant~$c$. In this case, we return the best input pair we found so far, i.e., $t_1^*$ and $t_2^*$, and the largest Lipschitz constant $\mathbf{Lip}(t_1^*,t_2)$ observed. Note that the returned constant is smaller than~$c$.
	
	In summary, the proposed method is an alternating optimisation scheme based on compass search. Basically, we start from the given $t_0$ to search for an image $t_1$ in a norm ball or hypercube, where the optimisation trajectory on the norm ball space is denoted as $S(t_0,\Delta(t_0))$) such that $Lip(t_0,t_1)>c$ (this step is symbolic execution); if we cannot find it, we modify the optimisation objective function by replacing $t_0$ with $t_1^*$ (the best concrete input found in this optimisation run) to initiate another optimisation trajectory on the space, i.e., $S(t_1^*,\Delta(t_0))$. This process is repeated until we have gradually covered the entire space $S(\Delta(t_0))$ of the norm ball.

\section{Test Oracle}
\label{sec:oracle}

%\youcheng{Why there are two oracles?}

We provide details about the validity checking performed for the generated test inputs (Line~9 of Algorithm~\ref{algo:testing}) and how the test suite is finally used to quantify the safety of the DNN.

\begin{definition}[Valid test input] \label{def:oracle1}
We are given a set $O$ of inputs for which we assume to have a correct classification (e.g., the training dataset). Given a real number $b$, a test input $t' \in \testsuites$ is said to be \emph{valid} if 
\begin{equation}
\exists t \in O: \distance{t-t'}{} \leq b \,.
\end{equation} 
\end{definition}
Intuitively, a test case $t$ is valid if it is close to some of the inputs for which we have a classification. Given a test input $t' \in \testsuites$, we write $O(t')$ for the input $t \in O$ that has the smallest distance to $t'$ among all inputs in $O$.  

To quantify the quality of the DNN using a test suite $\testsuites$, we use the following robustness criterion.
\begin{definition}[Robustness Oracle]\label{def:oracle2}
Given a set $O$ of classified inputs, a test case $t'$ passes the robustness oracle if 
\begin{equation}
\begin{array}{l}
%\exists (x,y) \in O: & \distance{x-x'}{} \leq b \\
\arg\max_{j} v[t']_{K,j} = \arg\max_{j} v[O(t')]_{K,j}
\end{array}
\end{equation} 
\end{definition}
Whenever we identify a test input $t'$ that fails to pass this oracle, then it serves as evidence that the DNN lacks robustness.

\section{Experimental Results}
\label{sec:exp}

We have implemented the concolic testing approach presented in this paper in a tool
we have named DeepConcolic\footnote{The implementation and all data in this section are available online at https://github.com/TrustAI/DeepConcolic}. We compare it with other tools for testing
DNNs. The experiments are run on a machine with 24 core Intel(R) Xeon(R) CPU E5-2620 v3 and 2.4\,GHz
and 125\,GB memory. We use a timeout of 12\,h. All coverage results are averaged over 10 runs or more.

\subsection{Comparison with DeepXplore}
\label{subsec:DeepConcolic+DeepXplore}

We now compare DeepConcolic and  Deep\-Xplore~\cite{deepxplore}
%\footnote{All 
%the data for DeepXplore are generated by using the software package available from 
%\url{https://github.com/peikexin9/deepxplore}.}
on DNNs obtained from the MNIST and CIFAR-10 datasets. We remark that DeepXplore has been applied to further datasets. 
%Unfortunately, the current DeepConcolic implementation does not support other DNN models
%used in DeepXplore.

%The DNN model architecture
%is given in Table~\ref{tbl:MNIST_Model} (Appendix~\ref{app:TestingResults}), and
%the configurations for DeepXplore are in Appendix~\ref{app:DeepXplore}.
 
%Results include the percentage of neuron coverage, together with some generated adversarial images,
%reported by the two software tools.

%\subsubsection{Neuron Cover Results}
For each tool, we start neuron cover testing from a randomly sampled image input.
Note that, since DeepXplore requires more than one DNN, we designate our trained DNN as the target model and utilise the other two default models provided by DeepXplore.
%To compare the two, we train a neural network with model architecture in Table~\ref{tbl:MNIST_Model} (Appendix~\ref{app:TestingResults}) and test accuracy 99.38\%, and we randomly choose an MNIST image 
%(index 6427) as the input seed. 
% Since DeepXplore requires multiple DNNs,
%the other two default DNNs provided by the tool are used. 
%
Table~\ref{tab:NC+DeepConcolic+DeepXplore} gives the neuron coverage obtained by the two tools.
We observe that DeepConcolic yields much higher neuron coverage than DeepXplore
in any of its three modes of operation (`light', `occlusion', and `blackout').
On the other hand, DeepXplore is much faster and terminates in seconds.

\begin{table}[th!]
    \caption{Neuron coverage of DeepConcolic and DeepXplore}
    \label{tab:NC+DeepConcolic+DeepXplore}
    \centering
    \def\arraystretch{1.2}
    \vspace{2mm}
    \scalebox{0.9}{
    \begin{tabular}{l|ccccc}
    \toprule
         & \multicolumn{2}{c}{\textbf{DeepConcolic}} & \multicolumn{3}{|c}{DeepXplore} \\ \cline{2-6}
         & $L_{\infty}$-norm & $L_{0}$-norm & \multicolumn{1}{|c}{light} & occlusion & blackout \\ \hline
        MNIST & 97.60\% & 95.91\% & \multicolumn{1}{|c}{80.77\%} & 82.68\% & 81.61\% \\
        CIFAR-10 & 84.98\% & 98.63\% & \multicolumn{1}{|c}{77.56\%} &81.48\% & 83.25\% \\
    \bottomrule
    \end{tabular}
    }
\end{table}

%\subsubsection{Adversarial Examples}

%For the same MNIST image, Figure~\ref{fig:MNIST+DeepXplore} presents 
%compares the original MNIST image (index 6427) with the
%adversarial examples generated from DeepConcolic ($L_{\infty}$-norm) and DeepXplore. Figure~\ref{fig:CIFAR-10+DeepXplore} 
%exhibits that of 
%presents adversarial examples of a random CIFAR-10 image (index 8661) from DeepConcolic ($L_0$-norm) and DeepXplore.

\begin{figure}[t!]
    \centering
    \includegraphics[width=1\linewidth]{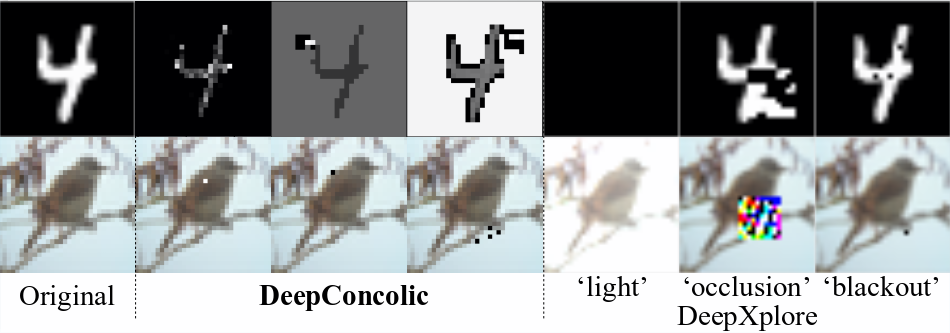}
    %\vspace{0.5pt}
    \caption{Adversarial images, with $L_{\infty}$-norm for MNIST (top row) and $L_0$-norm for CIFAR-10 (bottom row), generated by DeepConcolic and DeepXplore, the latter with image constraints `light', `occlusion', and `blackout'.}
    \label{fig:Adversaries}
\end{figure}

Figure~\ref{fig:Adversaries} presents several adversarial
examples found by DeepConcolic (with $L_{\infty}$-norm and $L_{0}$-norm) and DeepXplore.
Although DeepConcolic does not impose particular domain-specific constraints
on the original image as DeepXplore does, concolic testing generates images that resemble ``human perception''. For example, based on the $L_{\infty}$-norm, it produces adversarial examples (Figure~\ref{fig:Adversaries}, top row) that gradually reverse the black and white colours.
For the $L_0$-norm, DeepConcolic generates adversarial examples similar to those of DeepXplore under the `blackout' constraint, which is essentially pixel manipulation.

\subsection{Results for NC, SCC, and NBC}
\label{subsec:TestingResults}

We give the results obtained with DeepConcolic using the coverage criteria 
NC, SSC, and NBC. DeepConcolic starts NC testing with one single seed input.
For SSC and NBC, to improve the performance,
an initial set of 1000 images are sampled. Furthermore, we only test a subset
of the neurons for SSC and NBC. A distance upper
bound of 0.3 ($L_{\infty}$-norm) and 100 pixels ($L_0$-norm) is set up for
collecting adversarial examples.

\begin{figure}[ht!]
\centering
    %\begin{minipage}{0.27\linewidth}
%    \vspace{5pt}
	    \centering
	    \begin{minipage}{0.5\linewidth}
	        \centering
		    \includegraphics[width=1\linewidth]{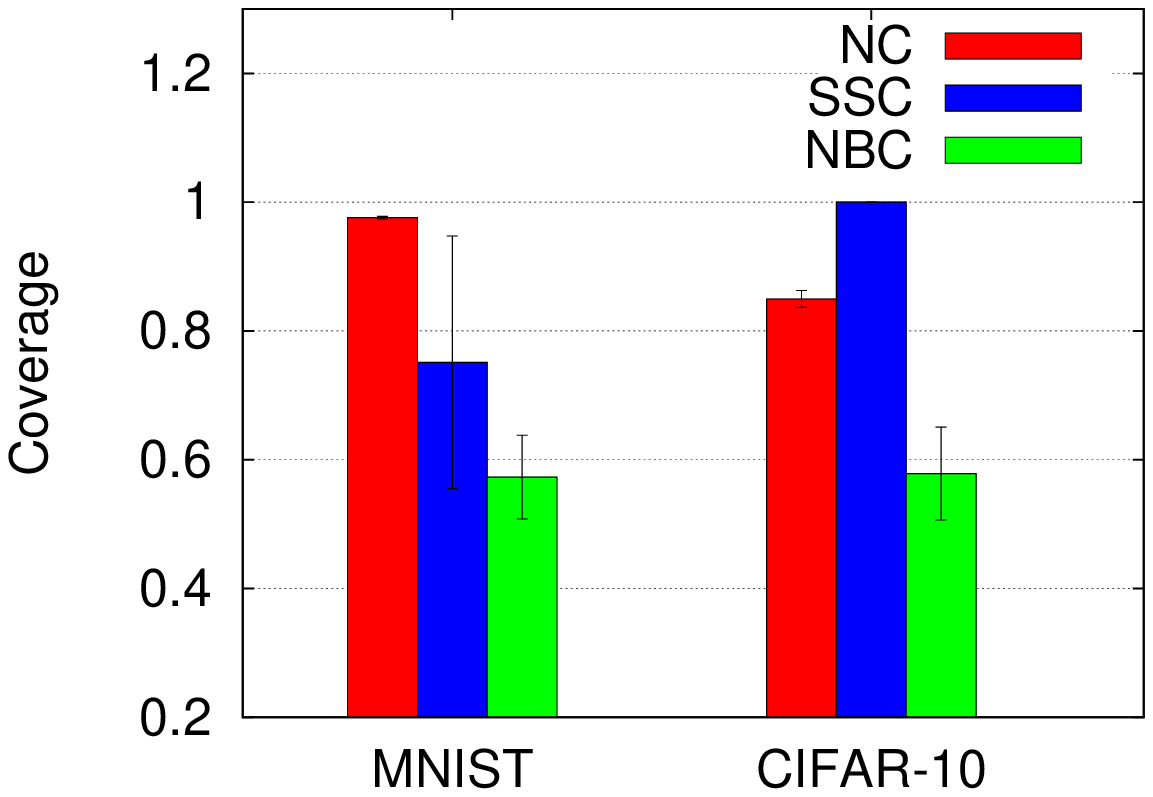}
		    \text{(a) $L_{\infty}$-norm}
	    \end{minipage}%
	    \begin{minipage}{0.5\linewidth}
		    \centering
		    \includegraphics[width=1\linewidth]{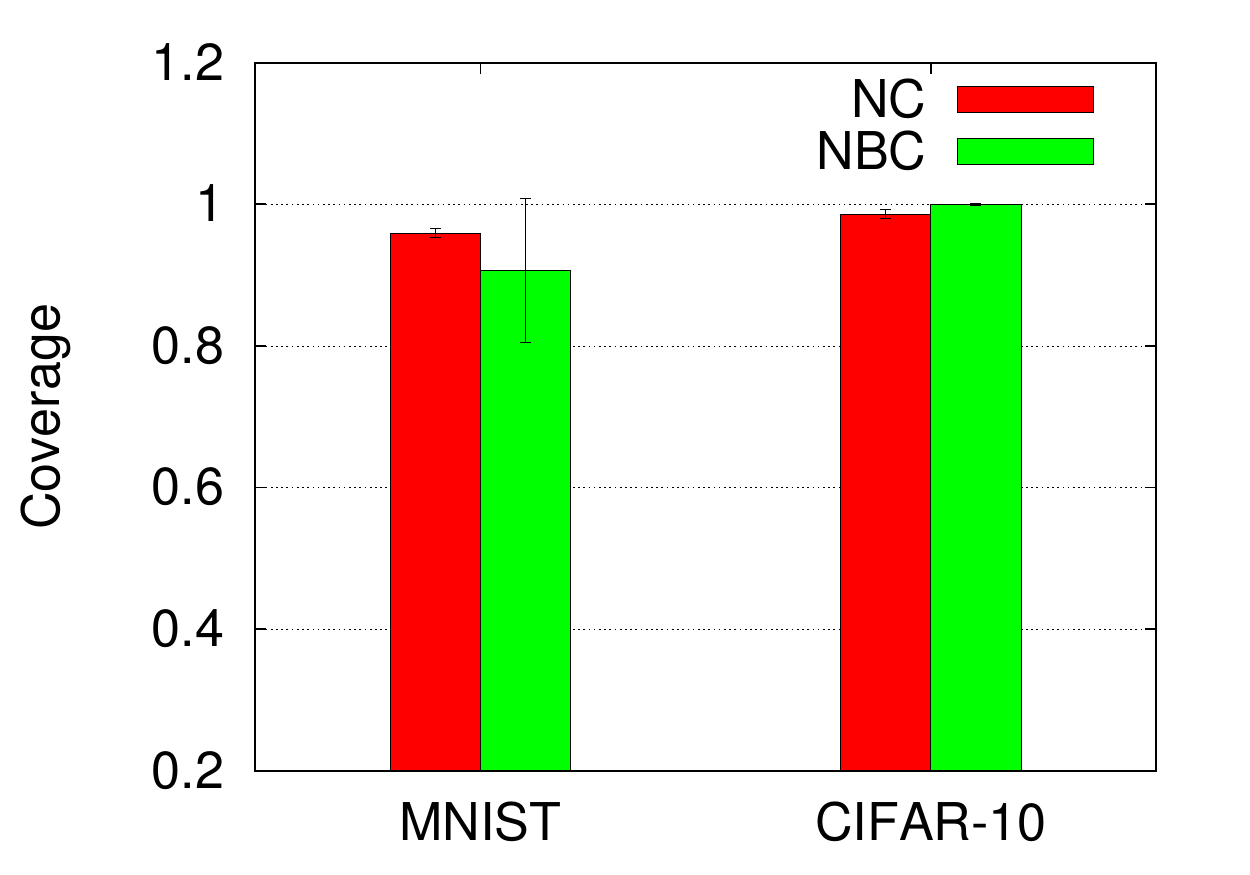}
		    \text{(b) $L_0$-norm}
	   \end{minipage}
	   \caption{Coverage results for different criteria}
	   \label{fig:coverage}

\end{figure}

The full coverage report, including the average coverage and standard derivation, is
given in Figure~\ref{fig:coverage}. Table~\ref{tab:cover-results} contains the adversarial example results. We have observed that the overhead for the symbolic analysis with global optimisation 
(Section~\ref{sec:symbol-optimisation}) is too high. Thus, the SSC result with $L_0$-norm is excluded.

\begin{figure}[t]
\centering
%    \vspace{5pt}
	    \centering
	    \begin{minipage}{0.3\linewidth}
	        \centering
		    \includegraphics[width=1\linewidth]{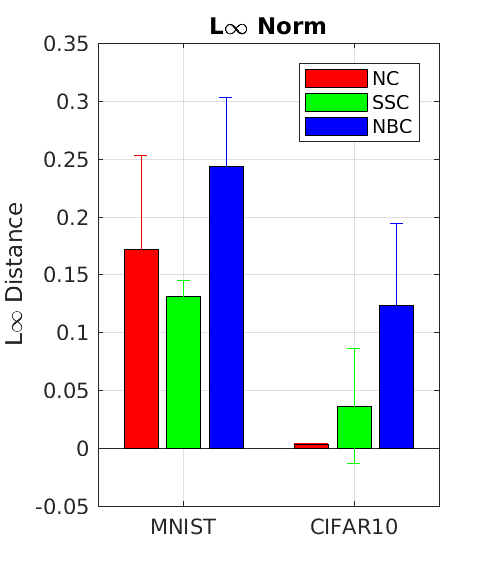}
		    \text{(a)}
	    \end{minipage}%
	    \begin{minipage}{0.2\linewidth}
		    \centering
		    \includegraphics[width=1\linewidth]{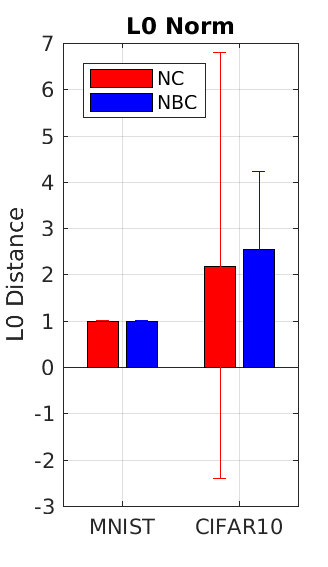}
		    \text{(b)}
	    \end{minipage}
	\vspace{10pt}
    \caption{(a) Distance of NC, SSC, and NBC on MINIST and CIFAR-10 datasets based on $L_\infty$ norm; (b)~Distance of NC and NBC on the two datasets based on $L_0$ norm.}
    \label{fig:Distance}
\end{figure}

\begin{figure}[t]
	\vspace{3pt}
	    \centering
	    \begin{minipage}{0.34\linewidth}
	        \centering
		    \includegraphics[width=1\linewidth]{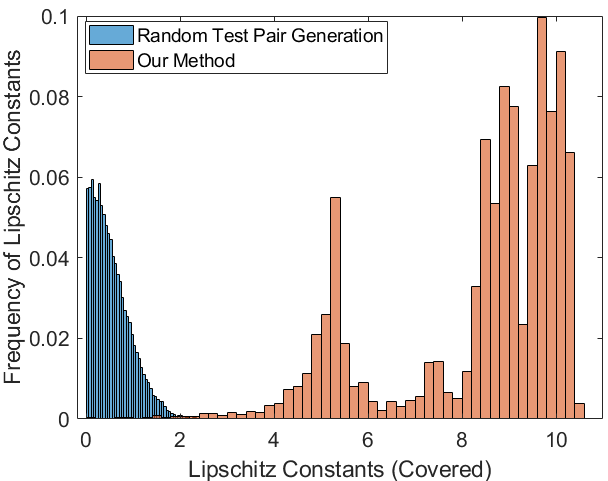}
		    \text{(a)}
	    \end{minipage}%
	    \begin{minipage}{0.33\linewidth}
		    \centering
		    \includegraphics[width=1.03\linewidth]{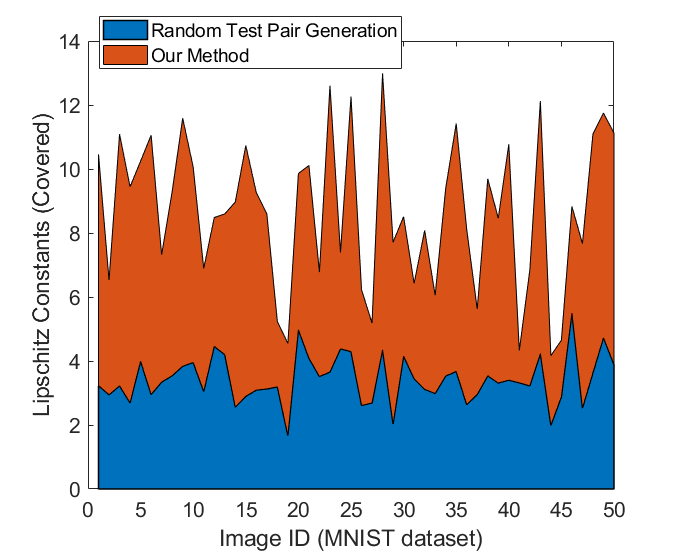}
		    \text{(b)}
	    \end{minipage}%
	    \begin{minipage}{0.33\linewidth}
		    \centering
		    \includegraphics[width=1.02\linewidth]{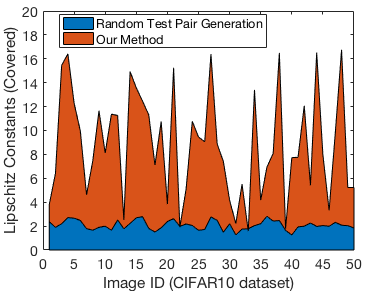}
		    \text{(c)}
	    \end{minipage}
	\vspace{1.5mm}
	\caption{(a) Lipschitz Constant Coverage generated by 1,000,000 randomly generated test pairs and our concolic testing method for input image-1 on MNIST DNN; (b) Lipschitz Constant Coverages generated by random testing and our method for 50 input images on MNIST DNN; (c) Lipschitz Constant Coverage generated by random testing and our method for 50 input images on CIFAR-10 DNN.}
	\label{fig1}
\end{figure}

\begin{table}[ht!]
    \centering
    \caption{Adversarial examples by test criteria, distance metrics, and DNN models}
    \label{tab:cover-results}
    \scalebox{0.55}{
    \begin{tabular}{l|cccccccc}
         \toprule
         & \multicolumn{4}{c}{\textbf{$L_{\infty}$-norm}} & \multicolumn{4}{|c}{\textbf{$L_{0}$-norm}} \\ \hline
         & \multicolumn{2}{c}{\textbf{MNIST}} & \multicolumn{2}{|c}{\textbf{CIFAR-10}}  & \multicolumn{2}{|c}{\textbf{MNIST}} & \multicolumn{2}{|c}{\textbf{CIFAR-10}} \\ \hline
         
         & \vtop{\hbox{\strut adversary \%}}
         & \vtop{\hbox{\strut minimum dist.}}
       
         & \multicolumn{1}{|c}{\vtop{\hbox{\strut adversary \%}}}
         & \vtop{\hbox{\strut minimum dist.}}
        
         & \multicolumn{1}{|c}{\vtop{\hbox{\strut adversary \%}}}
         & \vtop{\hbox{\strut minimum dist.}}
         & \multicolumn{1}{|c}{\vtop{\hbox{\strut adversary \%}}}
         & \vtop{\hbox{\strut minimum dist.}}\\ \hline
         \textbf{NC} &  13.93\% & 0.0039 &  \multicolumn{1}{|c}{0.79\%} & 0.0039  & \multicolumn{1}{|c}{0.53\%} & 1 & \multicolumn{1}{|c}{5.59\%} & 1 \\ 
         \textbf{SSC}  & 0.02\% & 0.1215  & \multicolumn{1}{|c}{0.36\%} & 0.0039  & \multicolumn{1}{|c}{--} & --  & \multicolumn{1}{|c}{--} & -- \\ 
         \textbf{NBC}  & 0.20\% & 0.0806  & \multicolumn{1}{|c}{7.71\%} & 0.0113  & \multicolumn{1}{|c}{0.09\%} & 1  & \multicolumn{1}{|c}{4.13\%} & 1 \\
         \bottomrule

    \end{tabular}
    }
\end{table}

Overall, DeepConcolic achieves high coverage and, using the robustness check (Definition~\ref{def:oracle2}), detects a significant number of
adversarial examples. However, coverage of corner-case activation values (i.e., NBC)
is limited.

Concolic testing is able to find adversarial examples
    with the minimum possible distance: that is, $\frac{1}{255}\approx 0.0039$ for the $L_{\infty}$ norm and $1$ pixel for the $L_0$ norm.
    Figure~\ref{fig:Distance} gives the average distance of adversarial examples (from one DeepConcolic run).
    %The effectiveness of a criterion varies when the DNN under test changes. For example,
    %subject to the $L_{\infty}$ norm, the NC seems to be more effective than SSC and NBC in the MNIST network with respect to the proportion of adversarial examples found; however, this is not the case when the CIFAR-10 network is tested. According to our results, different test criteria complement each other.
    %Remarkably, for the same CIFAR-10 network, many more adversarial examples are found for the NC when the $L_0$-norm is used.
    Remarkably, for the same network, the number of adversarial examples found with NC can vary substantially
    when the distance metric is changed.
    This observation suggests that, when designing coverage criteria for DNNs, they need to be examined using a variety of distance metrics.

\subsection{Results for Lipschitz Constant Testing}
\label{sec:LipConstTest}

%
%\subsubsection{Dataset} 
%We evaluate our 
%proposed 
%concolic testing method on 
%two 
%notable datasets - 
%MNIST and CIFAR-10 datasets.
%
%\subsubsection{DNN Structures} 
%two DNNs trained on MNIST and CIFAR-10 with 99.4\% and 76.6\% testing accuracy, respectively. Their details are %given in Appendix~\ref{app:LipConstTest}.

%\subsubsection{Experimental Setup} 
This section reports experimental results for the Lipschitz constant testing on DNNs.
We test Lipschitz constants ranging over $\{0.01:0.01:20\}$ on 50 MNIST images and %$\{0.01:0.01:20\}$ on 
50 CIFAR-10 images respectively. % chosen from testing datasets. 
Every image represents a subspace in $D_{L_1}$ and thus a requirement in Equation (\ref{eq:libr}). % We let $\Delta_{L_{\infty}} = 0.1$. The detailed parameter setup can be found in Appendix~\ref{app:LipConstTest}.

\subsubsection{Baseline Method} Since this paper is the first %research effort on testing 
to test Lipschitz constants of DNNs, %as far as we know there is no existing work that can be directly used as a baseline. Thus, 
we compare our method with random test case generation. For this specific test requirement, given a predefined Lipschitz constant $c$, 
%a network $\networks$, 
an input $t_0$ and the radius of norm ball (e.g., for $L_1$ and $L_2$ norms) or hypercube space (for $L_\infty$-norm) $\Delta$, we randomly generate two test pairs $t_1$ and $t_2$ that satisfy the space constraint (i.e., $||t_1 - t_0||_{D_2} 
\leq\Delta$ and $||t_2 - t_0||_{D_2} \leq \Delta$), and then check whether $\mathbf{Lip}(t_1,t_2) > c$ holds. We repeat the random 
%test cases 
generation until we find a satisfying test pair or the number of repetitions is larger than a predefined threshold. We set such threshold as $N_{rd} = 1,000,000$. Namely, if we randomly generate 1,000,000 test pairs and none of them can satisfy the Lipschitz constant requirement $>c$, we treat this test as a failure and return the largest Lipschitz constant found and the corresponding test pair; otherwise, we treat it as successful and return the satisfying test pair.

\subsubsection{Experimental Results}
Figure~\ref{fig1} (a) depicts the Lipschitz Constant Coverage generated by 1,000,000 random test pairs and our concolic test generation method for image-1 on MNIST DNNs. As we can see, even though we produce 1,000,000 test pairs by random test generation, the maximum Lipschitz converage reaches only 3.23 and most of the test pairs are in the range $[0.01,2]$. Our concolic method, on the other hand, can cover a Lipschitz range of $[0.01, 10.38]$, where most cases lie in $[3.5,10]$, which is poorly %hardly
covered by random test generation.

Figure~\ref{fig1} (b) and (c) compare the Lipschitz constant coverage of test pairs from the random method and the concolic method on both MNIST and CIFAR-10 models. Our method significantly outperforms random test case generation. We note that covering a large Lipschitz constant range for DNNs is a challenging problem since most image pairs (within a certain high-dimensional space) can produce small Lipschitz constants (such as 1 to 2). This explains the reason why randomly generated test pairs concentrate in a range of less than 3. However, for safety-critical applications such as self-driving cars, a DNN with a large Lipschitz constant essentially indicates it is more vulnerable to adversarial perturbations~\cite{RWSHKK2018,RHK2018}. As~a result, a test method that can cover larger Lipschitz constants provides a useful robustness indicator for a trained DNN. We argue that, for safety testing of DNNs, the concolic test method for Lipschitz constant coverage can complement existing methods to achieve significantly better coverage. 
%complements one important piece that is omitted by current DNN testing approaches. 

\section{Conclusions}
\label{sec:concl}

In this paper, we propose the first concolic testing method for 
DNNs. We implement it in a software tool and apply the tool to
evaluate the robustness of well-known DNNs. The generation of the
test inputs can be guided by a variety of coverage metrics, including
Lipschitz continuity. Our experimental results confirm that the
combination of concrete execution and symbolic analysis delivers
both coverage and automates the discovery of adversarial examples.

\subsubsection*{Acknowledgements}

This document is an overview of UK MOD (part) sponsored research and is released for informational purposes only. The contents of this document should not be interpreted as representing the views of the UK MOD, nor should it be assumed that they reflect any current or future UK MOD policy. The information contained in this document cannot supersede any statutory or contractual requirements or liabilities and is offered without prejudice or commitment.

\bibliographystyle{IEEEtran}
\bibliography{all.bib}

\end{document}